\documentclass[a4paper,12pt]{spieman}  
\usepackage{amsmath,amsfonts,amssymb}
\usepackage{graphicx}
\usepackage{setspace}
\usepackage{tocloft}
\usepackage[left, modulo]{lineno}

\title{Artificial intelligence as a gateway to scientific discovery: Uncovering features in retinal fundus images}
\author[a,b,*]{Parsa Delavari}
\author[a]{Gulcenur Ozturan}
\author[c]{Ozgur Yilmaz}
\author[a,b]{Ipek Oruc}
\affil[a]{University of British Columbia, Faculty of Medicine, Department of Ophthalmology and Visual Sciences, 2550 Willow St., Vancouver, BC, Canada, V5Z 0A6}
\affil[b]{University of British Columbia, Faculty of Medicine, Neuroscience, Djavad Mowafaghian Centre for Brain Health, 2215 Wesbrook Mall, Vancouver, BC, Canada, V6T 1Z3}
\affil[c]{University of British Columbia, Faculty of Science, Department of Mathematics, 1984 Mathematics Rd., Vancouver, BC, Canada, V6T 1Z2}

\cftpagenumbersoff{figure}
\cftpagenumbersoff{table} 
\begin{document} 
\maketitle

\begin{abstract}

\vspace{6pt}

\noindent\textbf{Purpose:} Convolutional neural networks can be trained to detect various conditions or patient traits based on retinal fundus photographs, some of which, such as the patient sex, are invisible to the expert human eye. Here we propose a methodology for explainable classification of fundus images to uncover the mechanism(s) by which CNNs successfully predict the labels. We used patient sex as a case study to validate our proposed methodology.

\vspace{3pt}

\noindent\textbf{Approach:} First, we used a set of 4746 fundus images, including training, validation and test partitions, to fine-tune a pre-trained CNN on the sex classification task. Next, we utilized deep learning explainability tools to hypothesize possible ways sex differences in the retina manifest. We measured numerous retinal properties relevant to our hypotheses through image segmentation to identify those significantly different between males and females. To tackle the multiple comparisons problem, we shortlisted the parameters by testing them on a set of 100 fundus images distinct from the images used for fine-tuning. Finally, we used an additional 400 images, not included in any previous set, to reveal significant sex differences in the retina. 

\vspace{3pt}

\noindent\textbf{Results:} We observed that the peripapillary area is darker in males compared to females ($p=.023, d=.243$). We also observed that males have richer retinal vasculature networks by showing a higher number of branches ($p=.016, d=.272$) and nodes ($p=.014, d=.299$) and a larger total length of branches ($p=.045, d=.206$) in the vessel graph. Also, vessels cover a greater area in the superior temporal quadrant of the retina in males compared to females ($p=0.048, d=.194$).

\vspace{3pt}

\noindent\textbf{Conclusions:} Our methodology reveals retinal features in fundus photographs that allow CNNs to predict traits currently unknown, but meaningful to experts.

\end{abstract}


{\noindent \footnotesize\textbf{*}Parsa Delavari,  \linkable{parsadlr@student.ubc.ca} }

\begin{spacing}{1}   

\section{Introduction}
\label{sect:intro}  

Deep neural networks (DNNs) are one of the most powerful machine learning models performing tasks previously considered to be exclusive to humans, such as visual object and face recognition, image segmentation, and natural language processing. Convolutional neural networks (CNNs) are a sub-type of deep learning networks specialized for image processing and classification. Relying on their superior performance, these models have found their way to almost every field of study, and medical imaging is no exception. CNNs are applied on various imaging modalities, such as computed tomography (CT), magnetic resonance imaging (MRI), and X-ray, to facilitate computer-aided detection (CAD) of abnormalities, and to assist clinicians in the process of diagnosis and management of diseases \cite{shen2017deep, suzuki2017overview}. 

Deep learning models have been widely used in retinal imaging modalities \cite{schmidt2018artificial, date2019applications}. Recently, CNNs have seen tremendous success in predicting eye diseases such as diabetic retinopathy (DR)~\cite{gulshan2016development}, age-related macular degeneration (AMD)~\cite{peng2019deepseenet}, and glaucoma~\cite{li2018efficacy} based on retinal fundus photographs. In addition to various signs of ocular diseases, features and traits not necessarily related to eye health are also visible to AI in this imaging modality. For example, Poplin et al. ~\cite{poplin2018prediction} successfully predicted cardiovascular risk factors, including patient's age, smoking status,  haemoglobin A1c (HbA1c), systolic blood pressure (SBP), diastolic blood pressure (DBP), and body mass index (BMI). Other systemic biomarkers, such as creatinine level, and body composition indices including muscle mass, height, and weight can also be predicted by CNNs based on fundoscopic images \cite{rim2020prediction}. Furthermore, the retina is considered a window to the brain, and it is the only part of the central nervous system that can be observed directly and non-invasively. The surface of the retina is covered by retinal ganglion cells (RGCs), a sub-type of neurons, and it shares many anatomical, physiological, and in turn, pathological properties with the brain. Several studies have shown the effects of Alzheimer's disease (AD) on the retina, bringing forward the potential for the development of new early diagnosis methodologies \cite{liao2018potential, sidiqi2020vivo, lee2020amyloid, mirzaei2020alzheimer, alber2020developing}. Recently, Tian et al.~\cite{tian2021modular} classified patients with Alzheimer's disease (AD) from healthy matched controls with more than 80\% accuracy based on fundus images and using machine learning methods. Their results indicate that AD symptoms in the retinal fundus photographs, which are, as of yet, invisible to the clinician eye, are present and detectable by AI in this imaging modality.

Despite the successful application of AI in a wide range of medical image classification tasks, the lack of transparency in its output decisions has been a roadblock in its widespread adoption in medicine \cite{ghassemi2021false}. Encapsulating the relationship between the input image and the output label is challenging in these so-called black-box models. Convolutional neural networks, in particular, consist of numerous layers and tens of millions of parameters. The inherent complexity in these family of models has led to the emergence of a separate field of study, ``deep learning interpretability", in search for human-comprehensible explainability tools for the underlying decision-making processes of the networks. Saliency maps, also known as heat maps or attention maps, are one of the most common interpretability tools for CNNs, and are used to highlight the image areas that have the greatest contribution to the model's decision \cite{simonyan2013deep, zeiler2014visualizing, selvaraju2017grad}. These saliency maps, however, offer only superficial information limited to the spatial distribution of regions used by the network, without a precise description of what features within the highlighted regions contributed to the model's decision. 

The other popular family of CNN explainability tools is feature visualization, which aims to uncover the visual features learned by the network, i.e., the preferred stimuli of a particular neuron, layer, or final label in the network \cite{zeiler2014visualizing, yosinski2015understanding, olah2017feature}. However, because of the complex nature of CNNs, a wide range of regularization techniques are needed to obtain meaningful and human-interpretable visualizations \cite{olah2017feature}. This problem is relevant for any classification tasks where human observers perform comparably with the models, but especially so for tasks in which CNNs show markedly superior performance or when human experts are not able to do the task at all. Due to these limitations, explainable AI has been referred to as a ``false hope" \cite{ghassemi2021false}. Therefore, especially in domains where the AI outperforms human observers, there is a gap between the superficial insight obtained from explainability tools, and the explicit knowledge of the, as yet unknown, diagnostic features that, in principle, contain useful information to support the classification task. Thus, in this study, we propose a novel and statistically sound methodology for using explainability to reverse engineer the latent information that is used by the AI. Importantly, we utilize explainability tools, not as a precise source of explicit information on how the model makes decisions, but instead, as a source of inspiration for forming exploratory hypotheses, which are then validated by statistical tests, eventually leading to scientific discoveries. 

The proposed methodology aims to reveal the features that are unknown to the human observer, that nevertheless allow a trained CNN to successfully accomplish a classification task. One such task is classification of patient's sex based on retinal fundus images. While invisible to the expert human eye (e.g. ophthalmologist), patient's sex is a trait that can be predicted successfully by CNNs based on fundus photographs \cite{poplin2018prediction, berk2022learning, korot2021predicting, molnar2020interpretable, ilanchezian2021interpretable}. Here, we use discrimination of male vs.~female eyes as a case study to validate the efficacy of our methodology. This particular classification task, although limited in clinical utility, serves as an ideal case for a proof-of-concept study considering the reliance of CNNs on large datasets. Indeed, patient's sex is a readily available label included in almost all medical imaging datasets with balanced samples of the two classes. 

Although ophthalmologists are not trained to recognize patient's sex in retinal imaging, it is conceivable that subtle differences exist between the male and female retinas. Studies using optical coherence tomography (OCT) have suggested retinal morphology differences between males and females, including retinal thickness and retinal nerve fiber layer thickness \cite{ooto2015effects, lamparter2018association}. In addition, sex differences in ocular blood flow have been reported, though few empirical studies have examined this issue (see Ref.~\citenum{schmidl2015gender} for a review). As the first study ever to show the ability of CNNs in predicting patient's sex based on fundus images, Poplin et. al. \cite{poplin2018prediction} used attention maps to highlight the regions that the trained model uses to make the predictions. They showed that for sex classification, the model mainly uses the optic disc and blood vessels, and these anatomical areas are highlighted in attention maps in 78\% and 71\% of the samples, respectively. Other studies have also suggested the optic disc, macula, and retinal vasculature as possible sources of sex differences relying on the saliency map results obtained from their trained models \cite{dieck2020factors, korot2021predicting}. Another study has used the BagNet model as an interpretable-by-design architecture to classify and explain sex based on retinal fundus images \cite{ilanchezian2021interpretable}. They demonstrated that the optic disc and macula provide evidence mostly for males and females, respectively. However, they stated that the specific features and sex differences within these anatomical areas contributing to the prediction are yet to be found. Inspired by AI's performance, Yamashita et. al. \cite{yamashita2020factors} compared numerous parameters and measurements available in color fundus images between men and women in order to find possible sex differences. They found multiple significant differences regarding the peripapillary area, optic disc, and retinal vessels. Combined, the statistically detected features achieved a classification accuracy of 77.9\%. Although this is an improvement compared to chance level, there is still a large gap between the achieved accuracy of this study and that of the best CNN results. Importantly, multiple parameters (about 40) were tested using the same dataset without accounting for multiple comparisons problem that can lead to false positive results. Moreover, as these parameters were defined based on clinical knowledge, it is unclear if AI uses the same factors in fundus images. The present study is the first of its kind that searches for specific and clear sex differences visible in retinal fundus images that are hypothesized solely based on deep learning and interpretation techniques.

\section{An overview of the proposed novel methodology}

By taking patient's sex as a case study, the proposed methodological pipeline consists
of three phases: (i) classifying sex by training a CNN, (ii) forming exploratory hypotheses using explainability tools, and (iii) testing hypotheses. In the first phase, we train a CNN model on sex classification based on fundus images, and the model's generalization performance is assessed to ensure the task is learned properly. In the second phase, deep learning interpretation techniques are utilized as a source of observation and inspiration to hypothesize a list of possible sex differences in the retina. Because of the discussed ambiguity in interpretation of AI explainability tools, at this stage we liberally list any feature that appears to be used by the model in forming these exploratory hypotheses. The resulting, potentially large number of hypotheses introduces a multiple comparisons problem. This issue arises when multiple statistical tests are applied simultaneously, leading to an increased probability of false positive outcomes. At the 0.05 significance level, one out of every 20 tests applied simultaneously is expected to yield significant results due to chance alone. On the other hand, being too stringent with statistical testing might result in missing potentially subtle sex-differences. To balance these two considerations, we utilize two separate and mutually-exclusive sets: a smaller, "exploration" set, is used to short-list the exploratory hypotheses through statistical tests without controlling for multiple comparisons. In the third phase, the selected hypotheses are tested a second time on the other, larger "verification" set, this time controlled for multiple comparisons.

\section{Methods}
\subsection{Overview}
In this work, we describe a novel methodological pipeline that leverages deep learning to reveal retinal features not currently part of diagnostician's toolkit within the standard of care. This proposed pipeline is composed of three phases. In \textit{Phase 1}, we train a CNN model to classify retinal fundus images based on a given binary label. Here, we have used patient sex (male vs.~female) as a case study, though the methodology can be extended to other traits and diagnostic labels. The set of fundoscopic images used to train and evaluate this model (\textit{the CNN Development Set}) is described in detail in Section~\ref{Sec:FundData}. In Sections~\ref{sec:cnn}--\ref{sec:eval} we specify the CNN architecture as well as the details of the training and evaluation procedure. In \textit{Phase 2}, we utilize CNN explainability tools to generate some insight as to what regions, anatomical landmarks, retinal features and signs visible in the fundoscopic images contributed to the successful classification of the trained model. These explainability tools are described in detail in Section~\ref{sec:explain}. These observations are then used to enumerate a liberally construed list of exploratory hypotheses. These hypotheses are articulated as potential differences in various retinal parameters which were tested on the \textit{Exploration Set} (see Section~\ref{Sec:FundData} for a detailed description). The various techniques to extract and measure the retinal parameters have been described in Section~\ref{sec:segmentation}. Once the exploratory hypotheses are statistically evaluated based on the Exploration Set, a select set of candidate hypotheses, i.e., the ones that yielded significant differences in Phase 2 were then passed on to \textit{Phase 3} for a rigorous statistical verification on an independent set of fundoscopic images--the \textit{Verification Set} (described in detail in Section~\ref{Sec:FundData}). Phase 3  statistically re-evaluates and attempts to verify whether positive results obtained in Phase 2 are replicated when tested on the Verification Set. The details of this final analysis is described in Section~\ref{sec:phase3-stats}.

\subsection{Fundus Image Datasets}\label{Sec:FundData}
Our methodology relies on forming three mutually-disjoint datasets: the \textit{CNN Development Set} (which is further partitioned into training, validation and test sets, as commonly done in neural network training), the \textit{Exploration Set}, and the \textit{Verification Set}.

We used two different sources of retinal fundus images in this study. The first, ODIR \cite{odir} is a publicly available fundus dataset containing 7,000 annotated images from 3,500 patients with information regarding their age, sex, and pathological condition of each eye. In this dataset, samples are labeled for diabetic retinopathy, glaucoma, cataract, hypertension, AMD, myopia, and other diseases or abnormalities. Images are collected from different hospitals and medical centers and therefore captured by a variety of fundus cameras with different resolutions and angles of view. To prevent the model from using potential sex differences in the prevalence of diseases, we excluded all images with any eye disease or other abnormalities, as well as images with low quality, as identified by a trained ophthalmologist (Dr.~Ozturan). The resulting dataset consists of 3146 images from 1991 patients, all of which contributed to the CNN development set. 
The summary statistics are reported in Table~\ref{tab:odir_sum}. The second source for fundus images was the retinal imaging database of Vancouver General Hospital Ophthalmic Imaging Department, named ``VCH-source" from here on. It comprises a total of 2100 images from 1167 patients. These images are selected from individuals with healthy eyes, assessed by Dr.~Ozturan to exclude the eyes with signs of ocular disease or abnormality. 1600 images (of 800 patients) of the VCH source was labeled as the ``VCH phase-1" set, and contributed to the CNN development set. Table~\ref{tab:dovs_sum} represents the summary statistics of the VCH phase-1 dataset. In sum, a total of 4746 images (ODIR + VCH-phase 1) were placed in the CNN development set and thus used to train and assess the CNN model. Both ODIR and VCH phase-1 portions of the development set were randomly partitioned into training, validation, and test sets with approximately 70\%, 15\%, and 15\% of the aggregate sets, respectively. Care was taken to make sure all images of any given patient remained in the same partition. All images were cropped to get a square image with equal height and width by detecting the circular contour of the fundus images and placing it at the center of the square.

The remaining 500 images from the VCH source were used to populate the \emph{Exploration} set (100 images, phase-2) and the \emph{Verification} set (400 images, phase-3). In both sets, exactly half of the male and female images are taken from the right eye. Age did not differ between male and female patients in both sets (\textit{p $>$ .2}). The summary statistics are reported in Table~\ref{tab:vch_sum}.

\begin{table}[ht]
\caption[Characteristics of the ODIR dataset]{Characteristics of ODIR dataset. The number of images and patients along with their age in each subset and in total are reported.}
\label{tab:odir_sum}
\begin{center}       
\resizebox{0.75\textwidth}{!}{
\begin{tabular}{rrrrr} 
\hline
\hline
\multicolumn{5}{c}{\textbf{ODIR}} \\[5pt]
Characteristics & Train & Validation & Test & Total \\[2pt]
\hline 
Number of Images & 2186 & 480 & 480 & 3146 \\
 Female & 972 & 222 & 220 & 1414 \\
 Male & 1214 & 258 & 260 & 1732 \\
\hline 
Number of Patients & 1511 & 240 & 240 & 1991 \\
 Female & 690 & 111 & 110 & 911 \\
Male & 821 & 129 & 130 & 1080 \\
\hline 
Average Patient's Age (SD) & 57.67 (11.40) & 57.04 (11.01) & 55.83 (11.42) & 57.37 (11.38) \\
 Female & 59.16 (11.54) & 57.92 (10.98) & 57.58 (11.06) & 58.82 (11.43) \\
 Male & 56.42 (11.14) & 56.26 (10.98) & 54.35 (11.50) & 56.15 (11.18) \\
\hline 
\hline
\end{tabular}}
\end{center}
\end{table}

\begin{table}[ht]
\caption[Characteristics of the VCH phase-1 dataset]{Characteristics of the VCH phase-1 dataset. The number of images and patients along with their age in each subset and in total are reported.}
\label{tab:dovs_sum}
\begin{center}   
\resizebox{0.75\textwidth}{!}{
\begin{tabular}{rrrrr} 
\hline 
\hline
\multicolumn{5}{c}{\textbf{VCH phase-1}} \\[5pt]
Characteristics & Train & Validation & Test & Total \\[2pt]
\hline 
Number of Images & 1120 & 240 & 240 & 1600 \\
 Female & 560 & 120 & 120 & 800 \\
 Male & 560 & 120 & 120 & 800 \\
\hline 
Number of Patients & 560 & 120 & 120 & 800 \\
 Female & 280 & 60 & 60 & 400 \\
 Male & 280 & 60 & 60 & 400 \\
\hline 
Average Patient's Age (SD) & 82.19 (5.61) & 82.07 (5.38) & 82.11 (5.50) & 82.16 (5.56) \\
Female & 82.52 (5.79) & 83.25 (5.26) & 82.50 (5.63) & 82.63 (5.69) \\
 Male & 81.86 (5.40) & 80.90 (5.24) & 81.72 (5.34) & 81.70 (5.38) \\
\hline 
\hline
\end{tabular}}
\end{center}
\end{table}

\begin{table}[ht]
\caption[Characteristics of the VCH phase-2 and phase-3 datasets]{Characteristics of the VCH phase-2 and phase-3 datasets. The number of images and patients, along with their age in each subset and in total are reported.}
\label{tab:vch_sum}
\begin{center}    
\resizebox{0.65\textwidth}{!}{
\begin{tabular}{rrrr} 
\hline 
\hline
\multicolumn{4}{c}{\textbf{VCH phase-2 and phase-3}} \\[5pt]
Characteristics & Exploration & Verification & Total \\[2pt]
\hline
Number of Images & 100 & 400 & 500 \\
Female & 50 & 200 & 250 \\
Male & 50 & 200 & 250 \\
\hline
Number of Patients & 75 & 297 & 367 \\
 Female & 36 & 146 & 179 \\
Male & 39 & 151 & 188 \\
\hline
Average Patient's Age (SD) & 43.52 (12.46) & 43.00 (11.53) & 43.17 (11.73) \\
 Female & 45.14 (13.29) & 42.78 (11.23) & 43.31 (11.68) \\
Male & 42.03 (11.44) & 43.21 (11.80) & 43.05 (11.78) \\
\hline 
\hline
\end{tabular}}
\end{center}
\end{table}

The study was approved by the UBC Clinical Research Ethics Board and Vancouver Coastal Health Research Institute and the requirement of consent was waived (H21-02013).

\subsection{CNN Architecture}
\label{sec:cnn}
The model architecture used in this study is Inception-Resnet-v2 \cite{szegedy2017inception}. The model’s parameters were initialized by the pretrained model weights on the ImageNet dataset \cite{deng2009imagenet}. Since the pretrained model has 1000 output classes consistent with the ImageNet classification contest, the model’s classifier module, which is the last fully connected layer,  was replaced with a new randomly initialized fully connected layer containing 1536 inputs (the number of output features of InceptionResNet-v2 model) and two outputs corresponding to male and female classes.

\subsection{Training Procedure}
\label{sec:trainig}

We utilized a transfer learning approach followed by a fine-tuning step to train the network using the training subset of the CNN development set. During the first 20 epochs, the network’s weights were frozen while the new classifier layer was learning the task. This is a common technique to prevent the gradient calculated based on the initial random weights of the classifier layer from changing the network’s parameters in a direction that is not meaningful and not necessarily aligned with the task. At the conclusion of the first 20 epochs, as the classifier has learned the task to some extent, we unfroze the network’s weights and allowed them to change during the subsequent 100 epochs. Hyperparameters were tuned based on the validation performance and by trying various combinations. A summary of the hyperparameters used for training and evaluating the model can be found in Table~\ref{tab:hyperparameters}.

\subsection{Data Augmentation and Transforms}
\label{sec:augmentation}

To further improve the performance of the model, we took advantage of data augmentation and image transforms. During the training process all images were rotated by a random amount chosen uniformly from -10 to +10 degrees to prevent the network from memorizing image-label pairs. Furthermore, we utilized a novel idea, not used before in similar studies to the best of our knowledge, that can be applied to fundus image datasets specifically because of their nature: The left and right retinas are anatomical mirror-images, approximately symmetrical along vertical axis, leading to a large image-level change (left vs.~right) that is not related or informative to the sex classification task. We removed this image variance across the data set by horizontally flipping all images taken from the right-eye so they appear like a left-eye fundus image. This ``horizontal flipping'' transform removes part of the image variance across the dataset that is known a priori to be irrelevant to the task and improves the model performance for sex classification. Arguably this is because horizontal flipping allows the model to expect the same anatomical parts of the retina in nearly same locations of the input image (i.e., optic disc on the left side and fovea on the right side) and, in turn, learns the features more efficiently. 

\begin{table}[ht]
\renewcommand{\arraystretch}{1.2}
\caption[CNN Hyperparameters]{Summary of the hyperparameters used for training and evaluating the model.}
\label{tab:hyperparameters}
\begin{center}    
\resizebox{0.65\textwidth}{!}{
\begin{tabular}{rr} 
\hline 
\hline
\multicolumn{2}{c}{\textbf{\large Training Hyperparameters}} \\[3pt]
\multicolumn{2}{l}{\textbf{Optimizer}}\\[2pt]
\hline 
\textbf{method} & Adam  \\
\textbf{batch size} & 16 \\
\textbf{number of epochs} & 120 \\
\textbf{initial learning rate} & 0.0003 \\
\textbf{learning rate annealing (first 20 epochs)} & 0.5 every 5 epochs \\
\textbf{learning rate annealing (next 100 epochs)} & 0.5 every 20 epochs \\
\\
\multicolumn{2}{l}{\textbf{Criterion}}\\[2pt]
\hline 
\textbf{loss function} & binary cross-entropy \\
\textbf{class weights (Female, Male)} & (0.463, 0.537) \\
\\
\multicolumn{2}{l}{\textbf{Network}}\\[2pt]
\hline
\textbf{architecture} & Inception-Resnet-v2 \\
\textbf{input image resolution} & 299$\times$299 \\
\textbf{number of features (hidden layer)} & 1536 \\
\textbf{number of output classes} & 2 \\
\\
\multicolumn{2}{l}{\textbf{Training Transforms}}\\[2pt]
\hline 
\textbf{random rotation} & $\theta \sim Uniform(-10^{\circ}, +10^{\circ})$ \\
\textbf{resize} & 299$\times$299 \\
\\
\multicolumn{2}{l}{\textbf{Validation Transforms}}\\[2pt]
\hline 
\textbf{resize} & 299$\times$299 \\
\\
\hline 
\hline
\end{tabular}}
\end{center}
\end{table}

\subsection{Model Evaluation}
\label{sec:eval}
The validation subsets of the CNN development dataset was used for evaluating the Model during the training process and tuning the hyperparameters. At every training epoch, various performance metrics were calculated and recorded on both training and validation sets: area under the receiver operating characteristic curve (AUC) score, accuracy, hit rate, false alarm rate, and binary cross-entropy (BCE) loss. Once the training course is over, the epoch at which the highest validation AUC occurred is selected to report validation metrics. In addition, the Model’s weights from the same epoch were saved as “the best model’s weights”. Then, the best Model was reloaded to obtain the performance metrics separately on the two unseen test sets obtained from ODIR and VCH phase-1 to assess generalizability performance. 

We used non-parametric bootstrapping to evaluate the significance of the results. We generated B = 1000 bootstrap replicates of the test sets to obtain the confidence interval for each performance metric. The chance level calculated from the ratio of male images in each test set was then compared to the AUC confidence intervals. p-values were calculated based on the percentile rank of chance-level (50\%) performance in the bootstrap AUC distribution.

In order to maintain a benchmark performance for comparison, three models are trained independently using the identical procedure described in Section~\ref{sec:trainig}, namely, “Normal Model”, “Flip Model”, and “Random Model”. The Flip Model is trained with the horizontal flip of right-eye images as explained earlier in Section~\ref{sec:augmentation}, while this transform is not used for training the Normal Model. Random Model was trained on the same datasets with the only difference that male and female labels were randomly shuffled in advance to training. Performance metrics on the test sets are reported for the three models.


\subsection{CNN Explainability}
\label{sec:explain}
We used the Grad-CAM \cite{selvaraju2017grad} technique to generate saliency maps. In this method, input images were first normalized by the average and SD values of each color channel calculated based on the ImageNet dataset (see \ref{tab:vis_hyperparameters}), the data on which the pretrained models are trained. Each image was then fed to the network to complete the forward path needed for calculating the gradient during the backward process, and the predicted label was saved. The model's output was one-hot coded, i.e., the output corresponding to the predicted class was set to one and the other class to zero. Next, the Grad-CAM saliency map was generated by back-propagating the gradient of the predicted label to the last convolutional layer. In an independent process, the Guided Backpropagation map was also calculated by the deconvolutional network created as an inverse of the trained model (see Ref.~\citenum{zeiler2014visualizing} for details) and again setting the predicted class's output to one and the other class's output to zero. Finally, these two matrices were multiplied pixel-wise to form the Guided Grad-CAM saliency maps.

To visualize the preferred stimuli for male and female classes, we use the regularized activation maximization method \cite{yosinski2015understanding}. This technique allows the user to input a sample image and provides a transformed version that maximizes the response at some layer of the network. Here, we input both noise images as well as male and female fundoscopic images from the test sets.  Similarly with the saliency map method, input images are normalized before going through the optimization process, and the model's parameters are fixed. Stochastic Gradient Descent (SGD) was used as the method to maximize the activation of the output class of interest. We tried different combinations of hyperparameters in the implementation of Ref.~\citenum{uozbulak_pytorch_vis_2022} and selected the values that led to qualitatively more interpretable results. The hyperparameters used to generate the final results are summarized in Table \ref{tab:vis_hyperparameters}. 

We used the PyTorch implementation by Ref.~\citenum{uozbulak_pytorch_vis_2022} for both Guided Grad-CAM and regularized feature visualization. The main challenge in visualizing CNN models is their complexity; therefore, a simpler model was used in this phase. A VGG16 \cite{simonyan2014very} model was trained using the same procedure explained in Section~\ref{sec:trainig} and achieved test AUC scores of 0.66 and 0.73 on ODIR and VCH phase-1 datasets, respectively. This model was used for interpretation purposes as it resulted in better and more understandable visualizations because of its simpler architecture compared to Inception-Resnet-v2. 

\begin{table}[ht]
\renewcommand{\arraystretch}{1.3}
\caption[CNN interpretation hyperparameters]{Summary of the hyperparameters used for saliency map generation and feature visualization.}
\label{tab:vis_hyperparameters}
\begin{center}       

\resizebox{0.5\textwidth}{!}{\begin{tabular}{rr} 
\hline 
\multicolumn{2}{c}{\textbf{\large Interpretation Hyperparameters}} \\[3pt]
\multicolumn{2}{l}{\textbf{Image Normalization}}\\[2pt]
\hline 
average of color channels (RGB) & [0.485, 0.456, 0.406]  \\
SD of color channels (RGB) & [0.229, 0.224, 0.225] \\
\\
\multicolumn{2}{l}{\textbf{Feature Visualization}}\\[2pt]
\hline 
number of iterations & 100 \\
Gaussian blur radius & 1 \\
blur frequency & every 10 iterations \\
Gaussian blur radius & 1 \\
SGD weight decay & 0.0001 \\
learning rate & 35 \\
clipping value & 0.5 \\
\hline 
\end{tabular}}
\end{center}
\end{table}

\subsection{Measuring Retinal Parameters}
\label{sec:segmentation}
In order to test the exploratory hypotheses, a wide variety of retinal parameters were measured. Thus, we needed to segment the main anatomical parts of the retina, specifically, the optic disc, retinal vasculature, and the peripapillary area. The segmentation results were then used to quantitatively measure the variables of interest and statistically test the hypothesized sex differences.

\begin{figure}
\begin{center}
\begin{tabular}{ccc}
\includegraphics[height=1.6in]{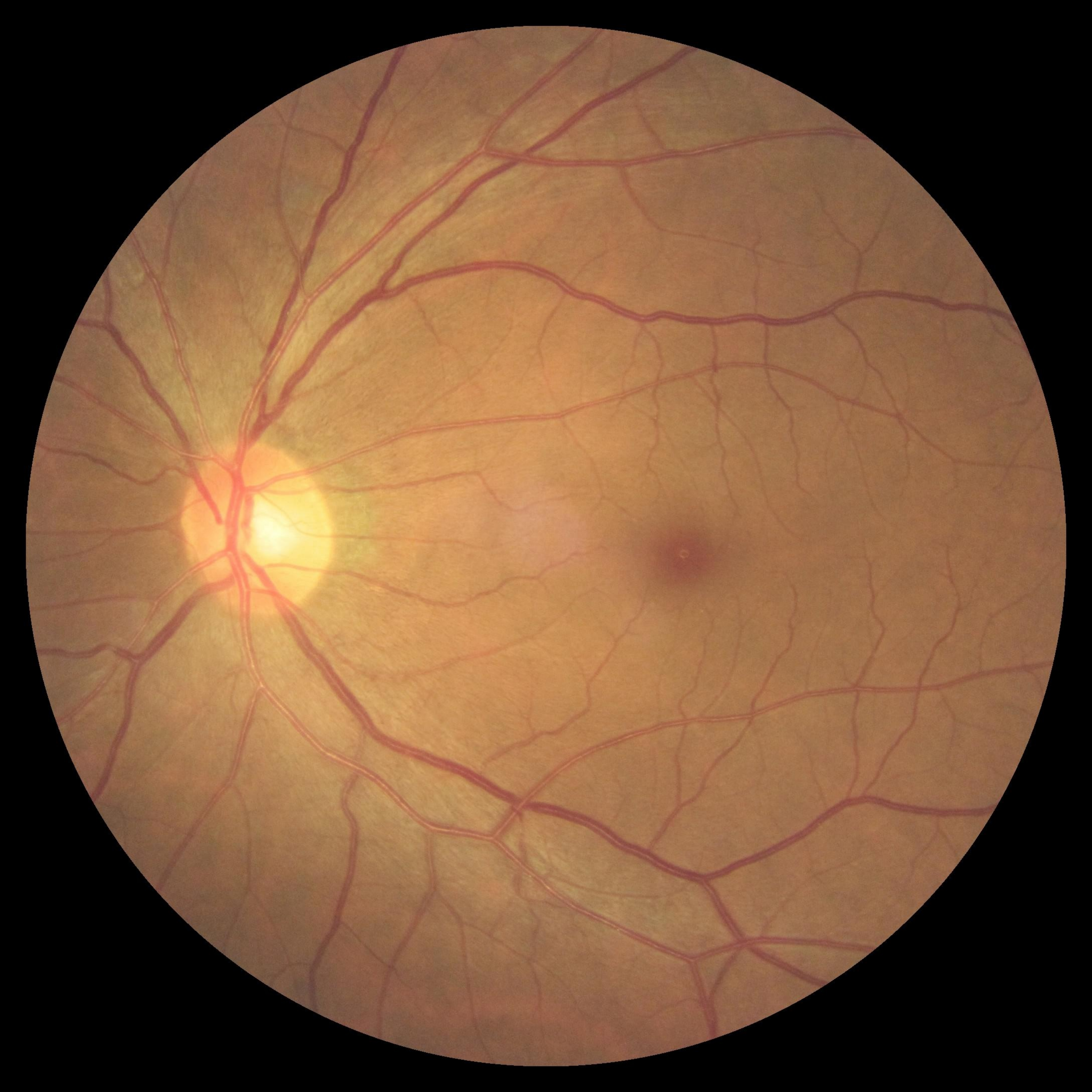} &  
\includegraphics[height=1.6in]{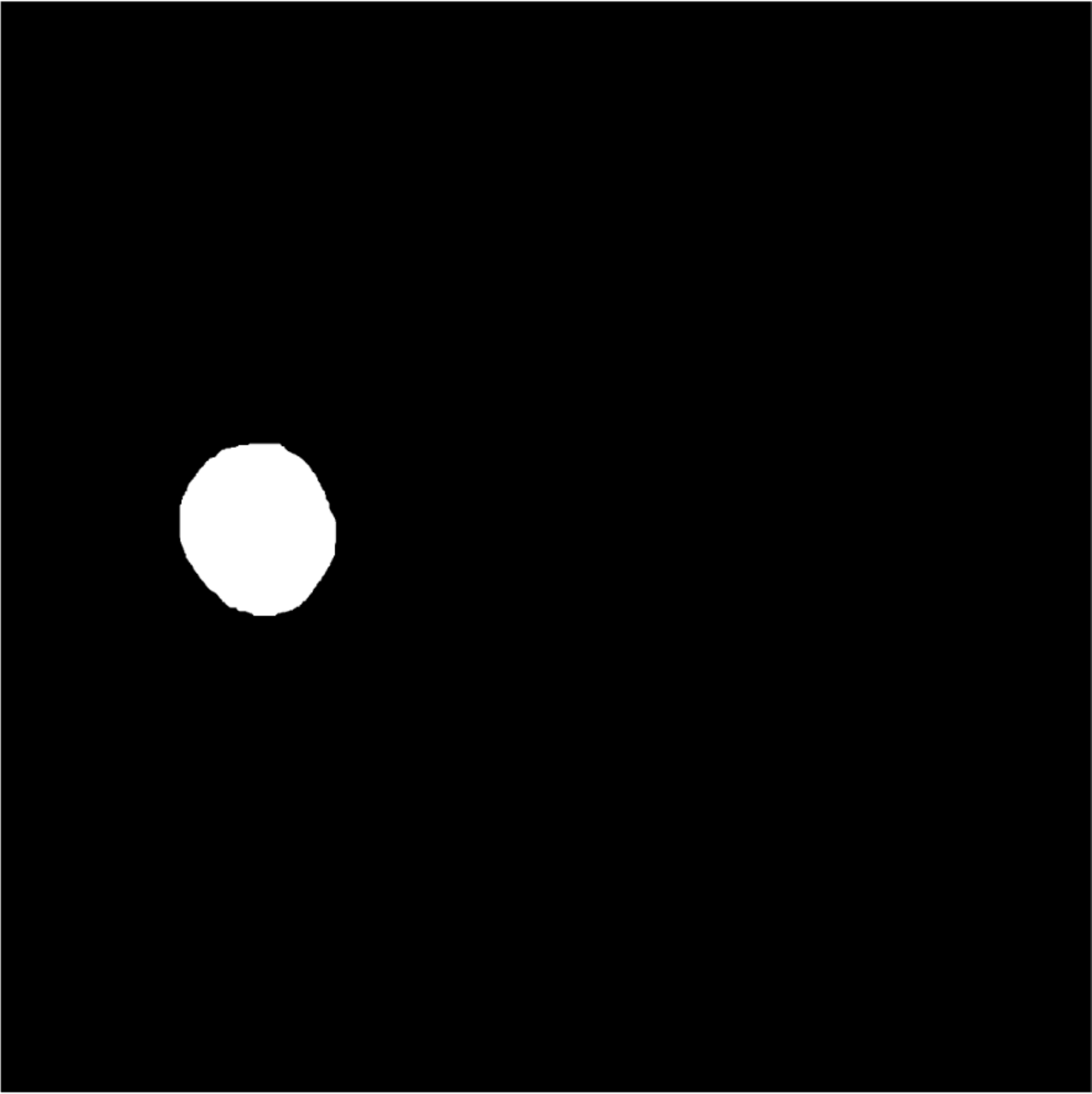} &
\includegraphics[height=1.6in]{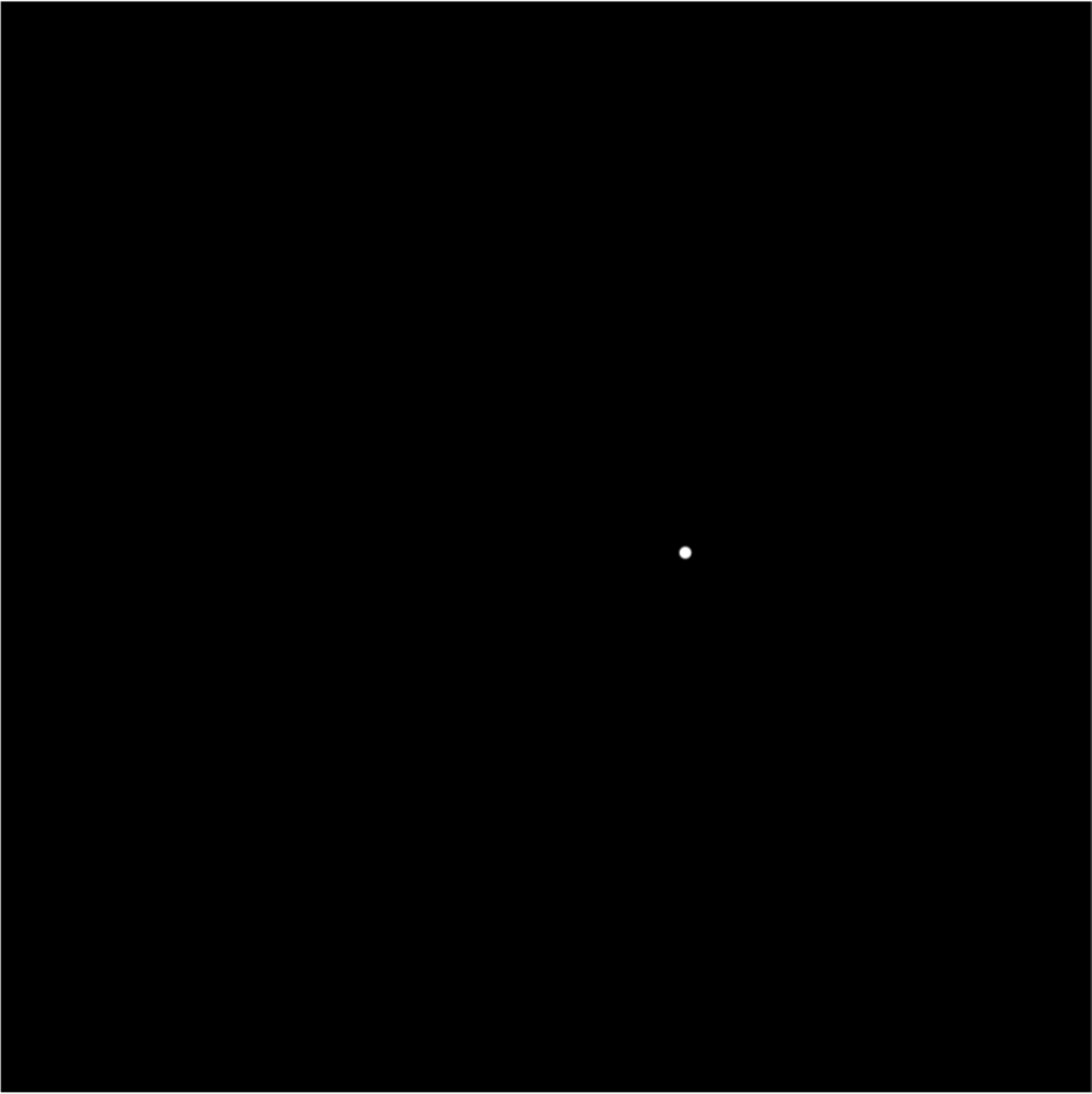} \\
(a) & (b) & (c)\\\\
\end{tabular}
\begin{tabular}{cc}
\includegraphics[height=1.6in]{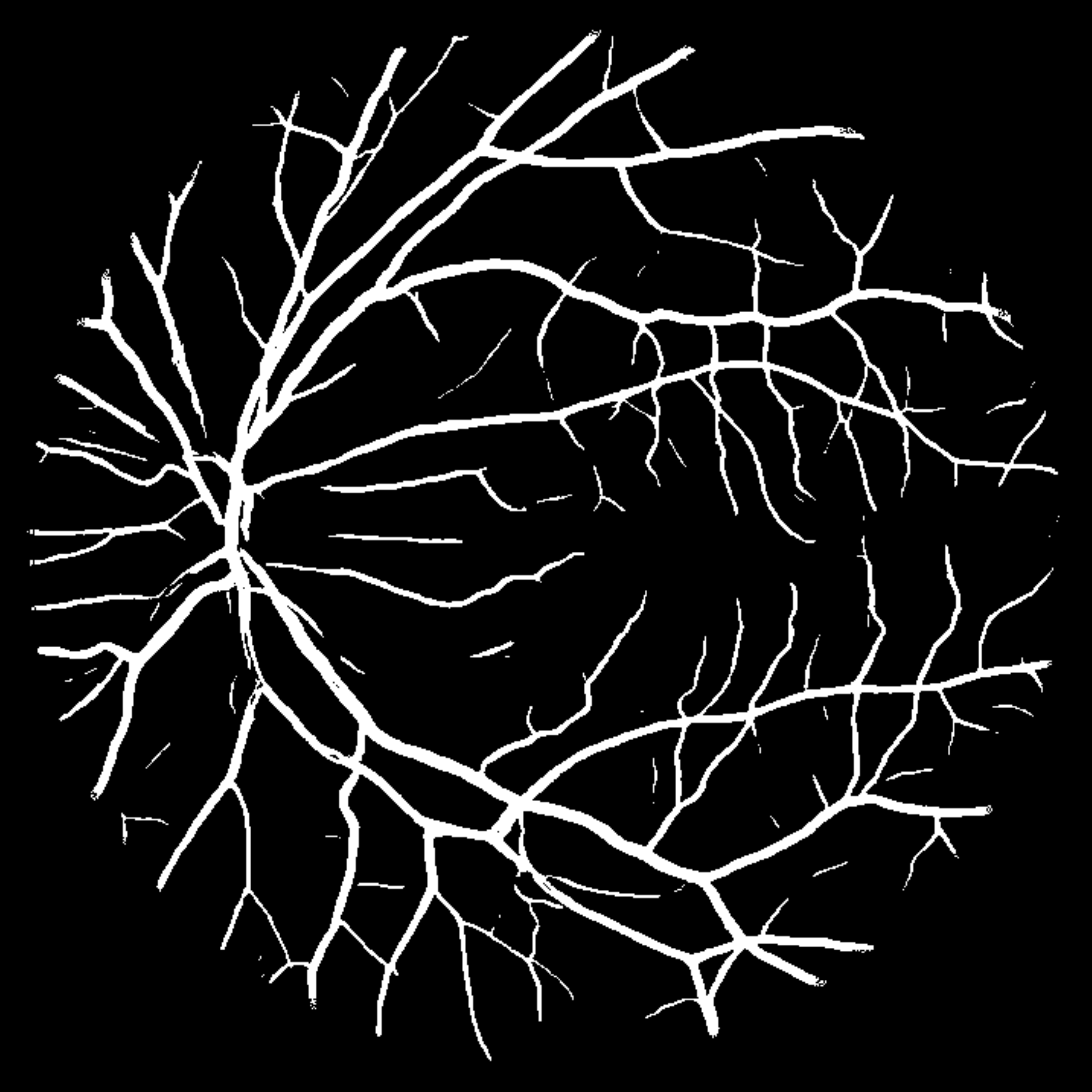} & 
\includegraphics[height=1.6in]{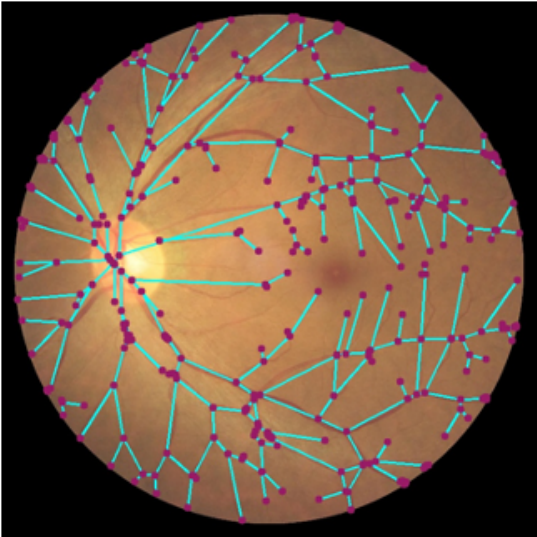} \\
(d) & (e) \\
\end{tabular}
\end{center}
\caption[Sample optic disc and fovea masks]{A sample fundus image (a) along with its binary optic disc mask (b), fovea (c), and vessel mask (d). An illustration of the vessel graph extracted based on binary vessel mask is depicted in panel (e). Light blue lines and red dots represent the edges and nodes of the obtained vessel graph, respectively.} 
\label{fig:masks} 
\end{figure}

The optic disc and the fovea were segmented using the Sefexa software \cite{sefexa} under the supervision of an ophthalmologist (Dr.~Ozturan). A sample segmentation output is depicted in Fig.~\ref{fig:masks}. The optic disc mask covers the non-parametric oval-like shape of the optic disc, and the fovea mask is a small dot marking the approximate location of the fovea since a precise location cannot be determined in this imaging modality. To segment the vasculature, we trained the LadderNet model \cite{zhuang2018laddernet}, a deep learning architecture specialized for image segmentation, using the DRIVE dataset. The implementation by Ref.~\citenum{lee_zq_2021} was used for training the model on the DRIVE dataset \cite{staal:2004-855} and then applying the trained model to the datasets used in this study. See the supplementary materials for the details of the vessel segmentation procedure and the DRIVE dataset. 

Based on the segmentation masks, multiple measurements were made to statistically test the hypotheses derived from CNN interpretation results. To characterize the optic disc (OD), we measured the area, sharpness of the edge, and brightness. The area covered by the vessels in different parts of the retina, as well as the number of nodes, number of branches, and the total length of branches in the vessel graph, were measured to characterize vasculature. Also, the radius of the foveal avascular zone (FAZ) was estimated. In order to account for the variation in the angular field of view, measurements were normalized by the distance between the fovea and the center of the optic disc, where appropriate. This is the same normalization method used by ophthalmologists; for instance, to assess the changes in optic disc size, they measure its diameter as a fraction of the distance between the fovea and optic disc.

In the definition of the measurements and variables, $G$ denotes the original fundus image with size $w \times h$ converted to gray-scale by using \texttt{cvtColor} function from OpenCV Python library. $G_{x,y}$ denotes the value of the pixel located at column x and row y of the gray-scale fundus image. In the binary masks (for the optic disc, vessels, and field of view (FOV)), each pixel can take the value of either zero (not annotated as the region of interest) or one (annotated as the region of interest). $OD_{x,y}$, $V_{x,y}$, and $FOV_{x,y}$ represent the value of pixel (x,y) of the optic disc mask, the vessel mask, and the FOV mask respectively.
\\
\noindent\textbf{Center of optic disc}\\
The optic disc center is calculated as the average coordinates of pixels included in the optic disc mask:
\[
x_\textrm{OD} = \dfrac{\sum_{x,y} \, x \, OD_{x,y}}{\sum_{x,y} \, OD_{x,y}}\:, \quad y_\textrm{OD} = \dfrac{\sum_{x,y} \, y \, OD_{x,y}}{\sum_{x,y} \, OD_{x,y}}
\]

\noindent\textbf{Distance between optic disc and fovea}\\
This distance is calculated to normalize length and area measurements to account for any differences in the angular view of the fundus images.
\[
d_\textrm{OD-F} = \sqrt{(x_\textrm{OD} - x_F)^2 + (y_\textrm{OD} - y_F)^2}
\]

\noindent\textbf{Normalized area of optic disc}\\
The area of the optic disc is calculated as the number of non-zero pixels in the optic disc mask. The resulting area is then normalized by a factor of $d_\textrm{OD-F}^2$ as its dimensionality is $pixel$-squared.
\[
A_\textrm{OD} = \sum_{x,y} OD_{x,y}\;,  \quad \quad
\bar{A}_\textrm{OD} = \dfrac{A_\textrm{OD}}{d_\textrm{OD-F}^2}
\]

\noindent\textbf{Average normalized brightness of optic disc}\\
The gray-scale image is first normalized in terms of brightness by shifting its average to 0.5. 
\[
\bar{G} = G - \dfrac{\sum_{x,y} G_{x,y}}{w \times h} + 0.5 
\]
This normalized gray-scale image, along with the optic disc mask, is then used to calculate the average brightness of the optic disc (denoted as $B_\textrm{OD}$). 
\[
B_\textrm{OD} = \dfrac{\sum_{x,y} OD_{x,y} \, \bar{G}_{x,y}}{\sum_{x,y} OD_{x,y}}
\]

\noindent\textbf{Sharpness of optic disc edge}\\
The Sobel operator is applied on the optic disc mask to obtain the contour of the optic disc ($S_\textrm{OD}$). Since the mask is binary, the Sobel operator returns a one-pixel wide contour, which is passed through a Gaussian blur filter (of size $7\times7$) to obtain a wider mask around the optic disc edge (denoted as $EM$). This resulting edge mask has its highest values on the exact edge of the optic disc and decreases as the pixels get farther from the optic disc edge.

\[
S_{\textrm{OD}} = \texttt{Sobel}(OD)
\]
\[
EM = \texttt{GaussianBlur}(S_{\textrm{OD}})
\]
Then the gray-scale image is passed through the Sobel operator to obtain the derivative of the image (denoted as $S$) and is averaged by the edge mask. 
\[
S = \texttt{Sobel}(\bar{G})
\]
\[
Sharpness_\textrm{OD} = \dfrac{\sum_{x,y} EM_{x,y} \; S_{x,y}}{\sum_{x,y} EM_{x,y}}
\]

\noindent\textbf{Brightness of peripapillary area}\\
First, the peripapillary mask is calculated by subtracting the optic disc mask from a circular mask with a radius of $1.4\,R_\textrm{OD}$ centered at the optic disc center. The resulting peripapillary binary mask denoted as $P$, is a ring-shaped mask with an average width of $0.4\,R_\textrm{OD}$ around the optic disc. 
\[
Brightness_{\textrm{peripapillary}} = \dfrac{\sum_{x,y} P_{x,y} \, \bar{G}_{x,y}}{\sum_{x,y} P_{x,y}}
\]

\noindent\textbf{Vessel coverage}\\
In addition to the entire fundus, we aimed to measure vessel coverage ($VC$) separately for different quadrants of the retina, namely, superior temporal (ST), superior nasal (SN), inferior temporal (IT), and inferior nasal (IN). Therefore, images are first aligned and rotated with respect to the center of the optic disc in a way that the line connecting the center of the optic disc to the fovea lies exactly horizontally. Also, the optic disc mask, fovea location, FOV mask, and the vessel mask are rotated accordingly and denoted with $r$ superscript.
\[
(VC)_{\textrm{entire fundus}} = \dfrac{\sum_{x,y} FOV_{x,y} \, V_{x,y}}{\sum_{x,y} FOV_{x,y}}
\]
For the right eye we have: 
\[
(VC)_{\textrm{ST}} = \dfrac{\sum_{x<x_c,y<y_c} FOV_{x,y}^r \, V_{x,y}^r}{\sum_{x<x_c,y<y_c} FOV_{x,y}^r}
\]
\[
(VC)_{\textrm{SN}} = \dfrac{\sum_{x>x_c,y<y_c} FOV_{x,y}^r \, V_{x,y}^r}{\sum_{x>x_c,y<y_c} FOV_{x,y}^r}
\]
\[
(VC)_\textrm{IT} = \dfrac{\sum_{x<x_c,y>y_c} FOV_{x,y}^r \, V_{x,y}^r}{\sum_{x<x_c,y>y_c} FOV_{x,y}^r}
\]
\[
(VC)_\textrm{IN} = \dfrac{\sum_{x>x_c,y>y_c} FOV_{x,y}^r \, V_{x,y}^r}{\sum_{x>x_c,y>y_c} FOV_{x,y}^r}
\]
And for the left eye we have: 
\[
(VC)_{\textrm{ST}} = \dfrac{\sum_{x>x_c,y<y_c} FOV_{x,y}^r \, V_{x,y}^r}{\sum_{x>x_c,y<y_c} FOV_{x,y}^r}
\]
\[
(VC)_{\textrm{SN}} = \dfrac{\sum_{x<x_c,y<y_c} FOV_{x,y}^r \, V_{x,y}^r}{\sum_{x<x_c,y<y_c} FOV_{x,y}^r}
\]
\[
(VC)_\textrm{IT} = \dfrac{\sum_{x>x_c,y>y_c} FOV_{x,y}^r \, V_{x,y}^r}{\sum_{x>x_c,y>y_c} FOV_{x,y}^r}
\]
\[
(VC)_\textrm{IN} = \dfrac{\sum_{x<x_c,y>y_c} FOV_{x,y}^r \, V_{x,y}^r}{\sum_{x<x_c,y>y_c} FOV_{x,y}^r}
\]
Macular vessel coverage is also measured using a circular mask centered at the fovea with the radius of $0.5\,d_\textrm{OD-F}$. The macula mask is denoted as $M$. 
\[
(VC)_\textrm{macula} = \dfrac{\sum_{x,y} M_{x,y} \, V_{x,y}}{\sum_{x<x_c,y>y_c} M_{x,y}}
\]

\noindent\textbf{FAZ radius}\\
The radius of FAZ is reported as the radius of the largest circle centered at the fovea, which does not contain any vessels according to the binary vessel mask. If we denote a circle mask centered at the fovea with radius $r$ as $C(r)$, we have:
\[
\begin{aligned}
R_\textrm{FAZ} \; = \; \max\ \quad & r\\
\textrm{s.t.} \quad & \sum_{x,y} C(r)_{x,y} \, V_{x,y} = 0\\
\\
\bar{R}_\textrm{FAZ} \; = \; \dfrac{R_\textrm{FAZ}}{d_\textrm{OD-F}} \; \; & 
\end{aligned}
\]

\noindent\textbf{Vessel graph properties}\\
To further analyze the structural properties of the retinal vasculature, the binary vessel masks were also translated to vessel graphs by a Python library named \texttt{Skan} designed for skeleton image analysis. The obtained vessel graph contains structural information such as details of branches and nodes (\ref{fig:masks}). Based on the output of the graph analysis, the number of branches, number of nodes, and the total length of branches for the vessel graph were calculated.

\subsection{Statistical Analysis}
\label{sec:phase3-stats}
A two-sample t-test (male eyes vs.~female eyes) has been performed for each of the retinal measurements. For the short-listed hypotheses (phase-3 "verification"), the p-values were then adjusted \cite{benjamini2009selective} using the Benjamini-Hochberg method \cite{benjamini1995controlling} to address the multiple comparisons problem.

\section{Results}


\subsection{Phase 1: Sex Classification Training Metrics}

\begin{figure}
    \centering
    \includegraphics[width=0.5\textwidth]{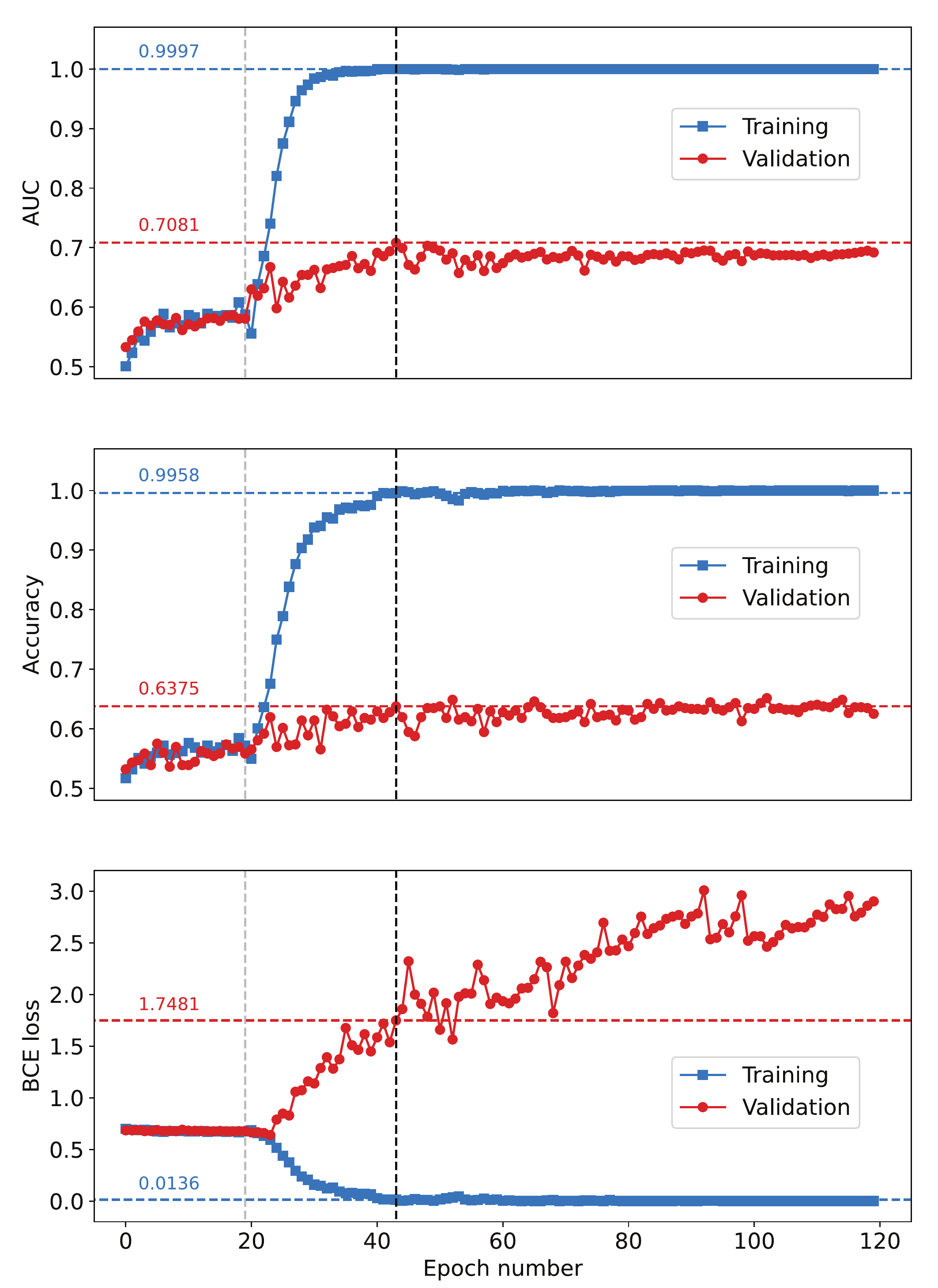}
    \caption[Training performance metrics]{From top to bottom, AUC, accuracy, and loss values are plotted across training epochs separately for training and validation sets. The dashed black line indicates the epoch at which the best model is selected, and the color lines represent the performance metrics of the best model.}
    \label{fig:training}   
\end{figure}

In order to track the training progress, the network’s performance was measured real-time over the epochs on both training and validation sets. Three evaluation metrics, namely AUC, accuracy, and BCE loss, are plotted separately for the training (blue) and validation (red) sets along with the learning rate in Fig.~\ref{fig:training}. The vertical black line indicates the epoch at which the highest validation AUC was achieved, and the model was saved for reporting metrics on the unseen test sets. The colored horizontal lines show the value of each metric at the best epoch on the training and validation sets. The vertical gray line shows the epoch at which the model is unfrozen, and the whole network is trained. The learning rate also resets to its initial value at this point. By looking at the patterns of changes in the performance metrics, we ensured that the model is learning the task properly and that the training course is running favorably. As shown in Fig.~\ref{fig:training}, AUC and accuracy, target indications of classification performance, improve over the training epochs on both training and validation sets. Although, the loss shows an increasing behavior on the test set a few epoch after unfreezing the network, this pattern has been attributed to the nature of BCE loss (see Ref.~\citenum{berk2022learning} for details).

\subsection{Phase 1: Sex Classification Test Results}

The three models were tested on the unseen ODIR and VCH phase-1 test partitions. To obtain the significance level of the results and calculate p-values, the performance metrics achieved by each model were compared to the chance level accuracy, which was 54.16\% and 50\% on ODIR and VCH phase-1 datasets, respectively (see Table~\ref{tab:cnn_res}).

\begin{table}[ht]
\renewcommand{\arraystretch}{1.3}
\caption[Sex classification resutls]{Test performance on the unseen test partitions of the ODIR and VCH phase-1 datasets. AUC and accuracy, along with their corresponding confidence intervals and p-values, are reported separately for the models on ODIR and DOVS test sets. * denotes significant p-values.}
\label{tab:cnn_res}
\begin{center} 
\resizebox{0.75\textwidth}{!}{
\begin{tabular}{c}
\begin{tabular}{ccccc}
    \hline 
    \multicolumn{5}{c}{(a) \textbf{ODIR}} \\[4pt]
    & AUC ($CI_\alpha$) & \textit{p-value} & Accuracy ($CI_\alpha$) & \textit{p-value} \\[2pt]
    \hline 
    Normal Model & 0.645 (0.597, 0.693) & \textit{$<$ .0001*} & 0.625 (0.579, 0.669) & \textit{$<$ .0001*} \\ 
    \textbf{Flip Model} & \textbf{0.676 (0.629, 0.724)} & \textbf{\textit{$<$ .0001*}} & \textbf{0.646 (0.602, 0.685)} & \textbf{\textit{$<$ .0001*}} \\
    Random Model & 0.480 (0.425, 0.533) & \textit{0.992} & 0.542 (0.498, 0.585) & 0.477 \\
    \hline 
\end{tabular}\\
\\
\begin{tabular}{ccccc}
    \hline 
    \multicolumn{5}{c}{(b) \textbf{VCH phase-1}} \\[4pt]
    & AUC ($CI_\alpha$) & \textit{p-value} & Accuracy ($CI_\alpha$) & \textit{p-value} \\[2pt]
    \hline 
    Normal Model & 0.760 (0.696, 0.818) & \textit{$<$ .0001*} & 0.692 (0.663, 0.746) & \textit{$<$ .0001*} \\ 
    \textbf{Flip Model} & \textbf{0.785 (0.727, 0.838)} & \textbf{\textit{$<$ .0001*}} & \textbf{0.721 (0.662, 0.775)} & \textbf{\textit{$<$ .0001*}} \\
    Random Model & 0.422 (0.351, 0.495) & \textit{0.979} & 0.496 (0.433, 0.562) & 0.476\\  
    \hline 
\end{tabular}
\end{tabular}}
\end{center}
\end{table}

On the unseen ODIR test partition, the AUC and accuracy achieved by the Normal Model were 0.645 and 0.625, respectively, both significantly greater than chance level ($p's<.001$). As anticipated, Flip Model performed slightly better than the Normal Model obtaining an AUC score of 0.676 ($p<.001$) and an accuracy of 0.646 ($p<.001$). The random model achieved 0.480 AUC ($p=.992$) and 0.542 ($p=.477$) accuracy, which are not significantly different from and very close to the chance level, as expected. The 95\% confidence intervals (CIs) are calculated based on non-parametric bootstrapping with $B=1000$ bootstrap resamples of the ODIR test set with size $N=480$. All metrics, p-values, and confidence intervals obtained from ODIR by the three models are shown in Table~\ref{tab:cnn_res}.a. The results imply that the network classifies patient sex significantly higher than chance level at an $\alpha = .05$ confidence level.

Based on the unseen VCH phase-1 test partition, the Normal Model obtained 0.760 AUC ($p<.001$) and 0.692 accuracy ($p<.001$). The Flip Model achieved higher performance on this dataset as well by an AUC of 0.785 ($p<.001$) and accuracy of 0.721 ($p<.001$). The Random model again performed similar to chance level with 0.422 AUC ($p=.979$) and 0.496 accuracy ($p=.476$). $B = 1000$ bootstrap resamples of the test set with $N=240$ sample size were used to calculate the confidence intervals. The AUC score, accuracy, p-values, and CI reported in Tabel~\ref{tab:cnn_res}.b indicates that the network has obtained a sex classification performance significantly higher than the chance level.

\subsection{Phase2: CNN Interpretation Results}

\subsubsection{Saliency Map}

\begin{figure}
\begin{center}
\begin{tabular}{cc}
\includegraphics[width=3in]{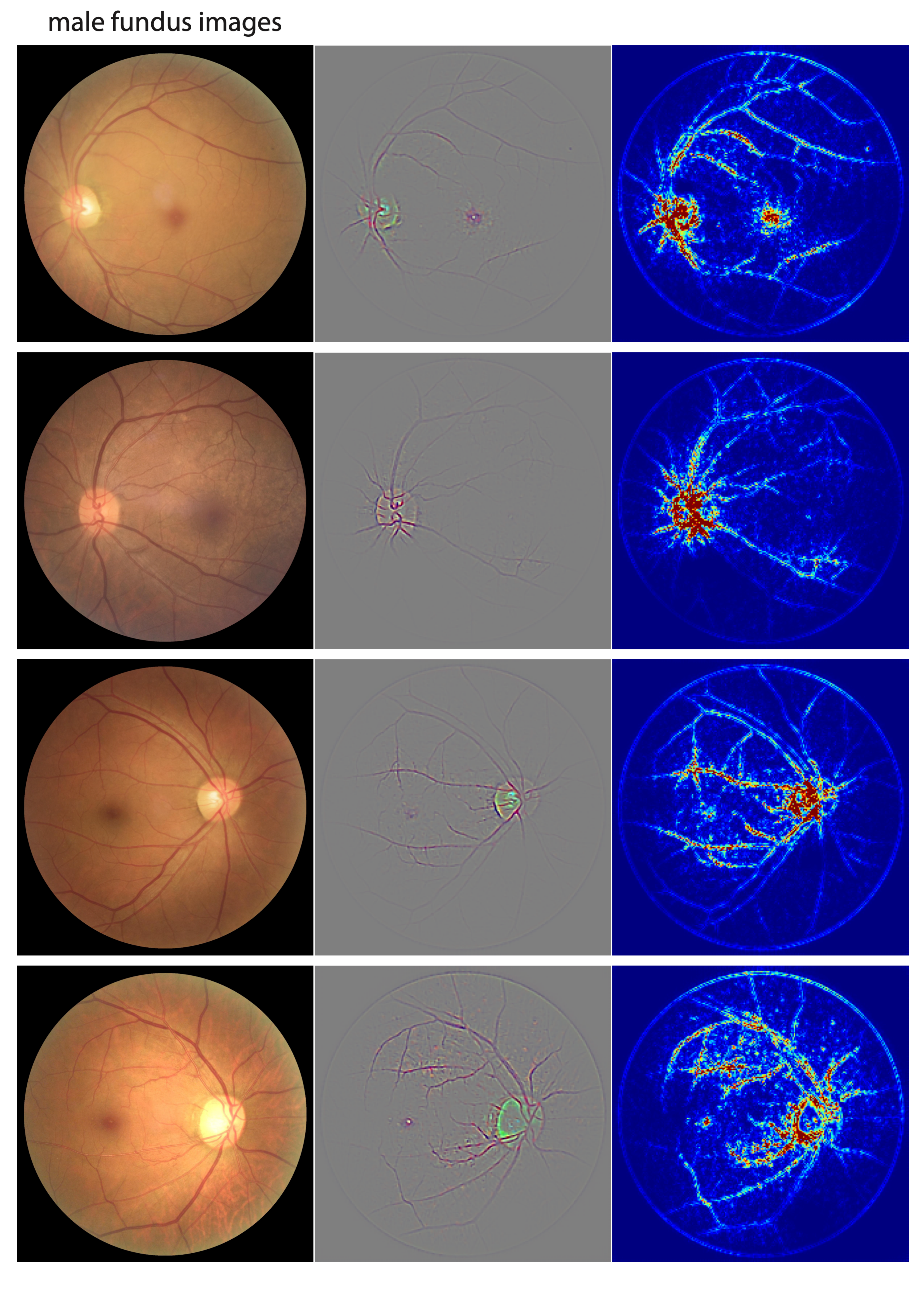} &  
\includegraphics[width=3in]{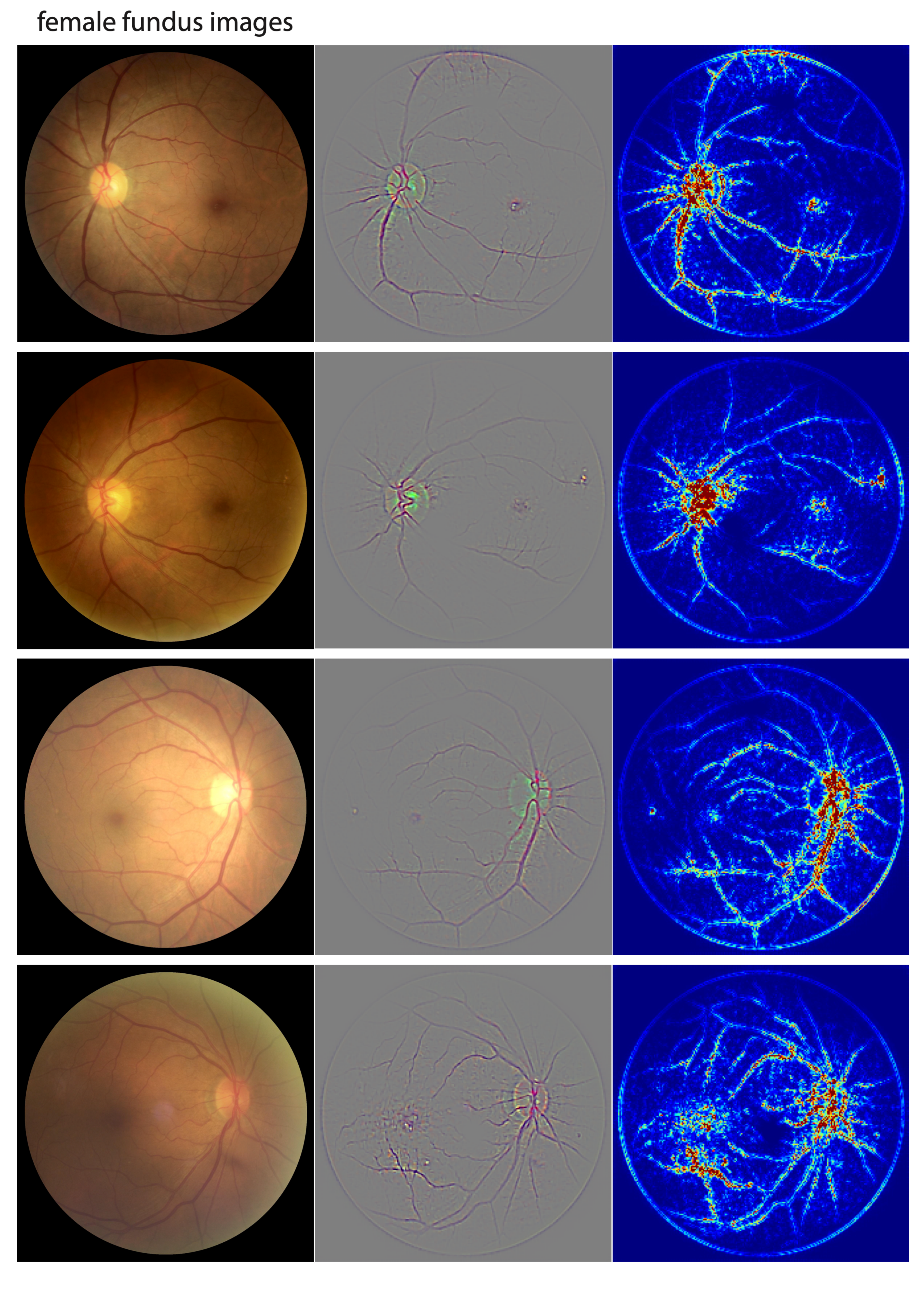} \\
(a) & (b) \\
\end{tabular}
\end{center}
\caption[Saliency map results]{Saliency map results of sample fundus images from four male (a) and four female patients (b). In each panel the original fundus image, the Guided Grad-CAM output (3-channel image), and its color-coded amplitude (single-channel image) are shown from left to right.}
\label{fig:saliency}  
\end{figure} 

Fig.~\ref{fig:saliency} depicts four sample saliency-map results for each of the male and female classes. The original fundus images fed to the model, the Guided Grad-CAM outputs, and the color-coded saliency maps are shown in the left, middle, and right panels, respectively. According to these illustrations, the network is mainly attending to the optic disc, retinal vasculature, and to some extent, the fovea. This pattern is consistent among different samples in both male and female groups, suggesting that the information needed for sex classification based on fundus photographs sits in these anatomical parts of the retina.

\subsubsection{Feature Visualization}

\begin{figure}
\begin{center}
\begin{tabular}{cc}
\includegraphics[width=3in]{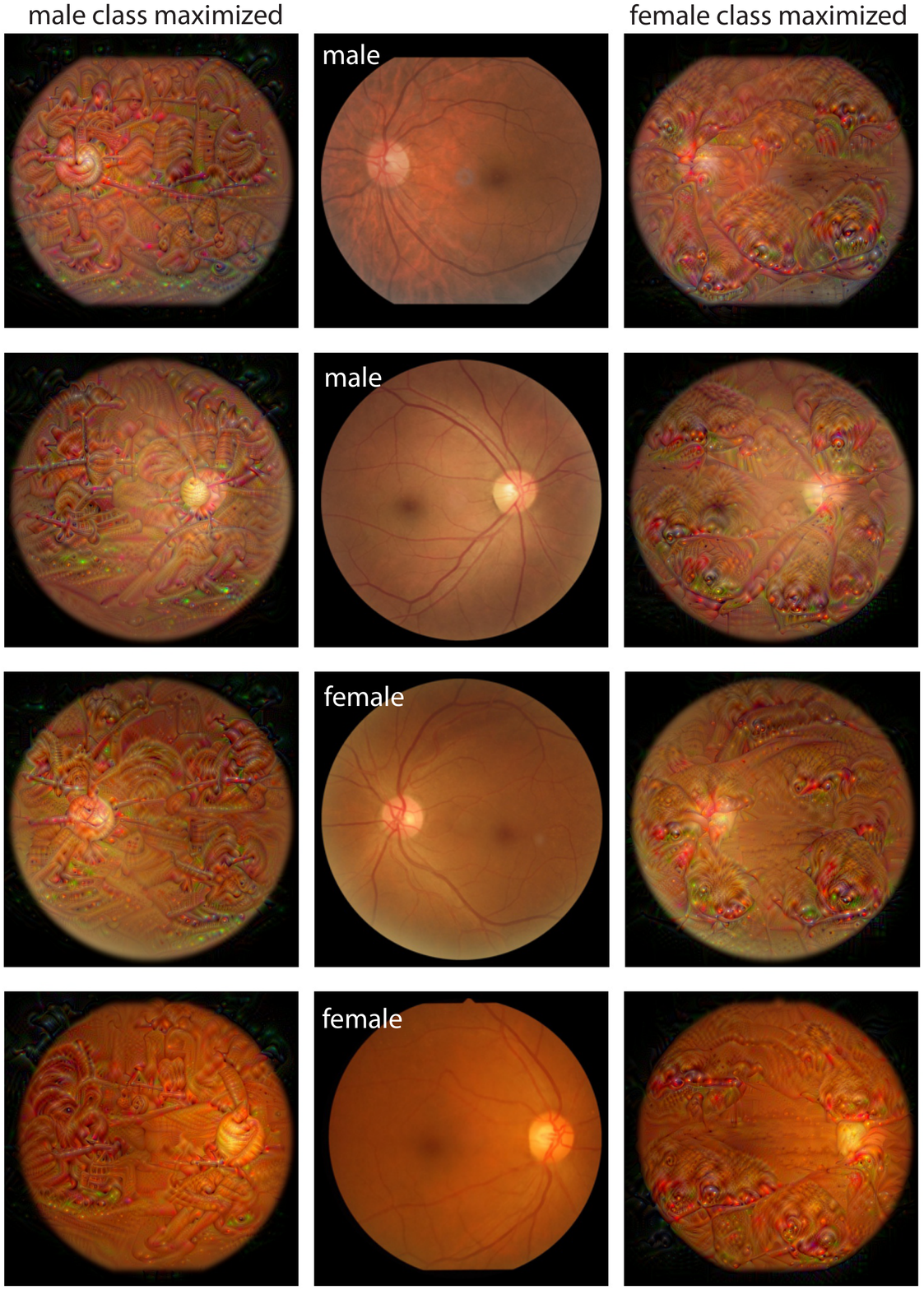}\ &  
\includegraphics[width=3in]{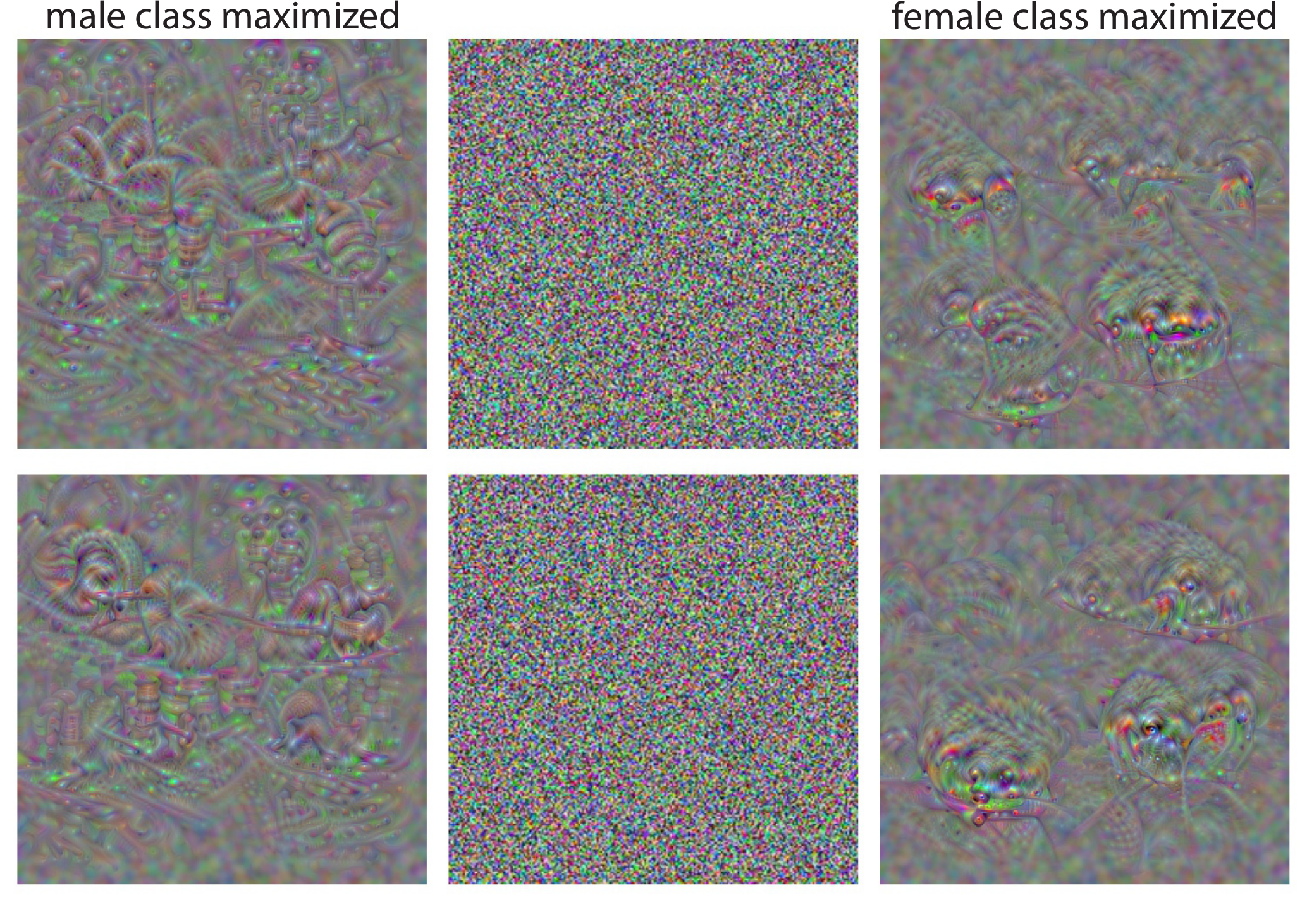} \\
(a) & (b) \\
\end{tabular}
\end{center}
\caption[Feature visualization results]{Feature visualization results for sample fundus images (a) and noise images (b). In panel (a), The top two rows of the middle column show original male images and the bottom two rows show original female images. The left and right columns represents feature visualizations for male and female classes respectively.} 
\label{fig:visulization} 
\end{figure}

Fig.~\ref{fig:visulization} shows four sample feature visualization results starting with a fundus image (middle column) as the initial image. The initial images are changed by the model to a more male-like and a more female-like fundus in the left and right columns, respectively. One of the clearest differences between synthetic male and female images appears to be that there are many vessel-like structures added to the original image for the male synthetic samples, whereas this pattern is not observed for the female synthetic samples. These added vessels are relatively thick and more similar to the main veins and arteries seen in the original fundus images. This suggests that from the model's perspective, males have more prominent and thicker retinal vasculature compared to females. Moreover, the optic disc seems to be more prominent in males as well, showing sharper edges and more contrast to the background. In contrast to males, the optic discs are somehow diminished in the female synthetic images. There is also a dark ring added around the optic disc in male synthetic samples, while it is not the case for female synthetic samples. Thus, based on the model's point of view, males have darker peripapillary areas than females. These are the most consistent patterns observed over many feature visualization outputs, which are present in both the left and the right eyes of synthesized images based on male and female fundus images (see Fig.~\ref{fig:visulization}a). In Fig.~\ref{fig:visulization}b, where the initial images are random noise, there are still thick vessels added to the male-like image and not to the female-like one. However, as expected, the changes regarding the optic disc are not observed in these results as there is no optic disc in the initial image.

\subsection{Phase 2: Exploratory Hypotheses}
\label{sec:exploratory_hypotheses}

Based on the observations on the saliency map and feature visualization results, possible sex differences in the retina were described as general exploratory hypotheses over three main themes: (1) retinal vessels are more prominent in males than females, (2) optic disc is more prominent in males than females, (3) the peripapillary area is darker in males than females. 

According to the two-sample t-test results, out of the 14 retinal parameters that were tested on the exploratory dataset, 9 were significantly different between males and females. The normalized area of optic disc is significantly larger in males. Female retinas show significantly brighter peripapillary areas compared to males. Regarding vessel properties, male eyes have higher vessel coverage in the superior temporal quadrant, inferior temporal quadrant, and macula. Also, the FAZ normalized radius is greater in males than in females. All three vessel graph properties, i.e., number of nodes, number of branches, and total length of branches, are significantly higher in males. On the other hand, we did not detect any significant sex differences in the normalized brightness of optic disc and the sharpness of optic disc edge. Also the vessel coverage in the superior nasal quadrant, inferior nasal quadrant, and macula did not show significant differences between the two groups. The average and standard deviation of all tested parameters for males and females, along with the p-values and effect sizes, are reported in Table~\ref{tab:sasndbox_results}.

\begin{table}[ht]
\renewcommand{\arraystretch}{1}
\caption[Exploratory results]{Statistical test results for sex differences in the defined measurements on the  exploration dataset.}
\label{tab:sasndbox_results}
\begin{center}       
\resizebox{0.9\textwidth}{!}{
\begin{tabular}{lcccr}
\hline 
\hline
Measurement & Male average (SD) & Females average (SD) & Effect size & \textit{p-value}\\
\hline
\hline
\\
\textbf{OD normalized area} & 0.117 (0.016) & 0.105 (0.020) & 0.712 & \textbf{\textit{$<$ .0001}}\\
OD normalized brightness & 0.808 (0.068) & 0.813 (0.083) & 0.072 & \textit{.3610}\\
OD edge sharpness & 0.198 (0.019) & 0.813 (0.023) & 0.223 & \textit{.1362}\\
\\
\textbf{Peripapillary area brightness} & 0.652 (0.052) & 0.670 (0.051) & 0.349 & \textbf{\textit{.0437}}\\
\\
\multicolumn{2}{l}{Vessel coverage}\\
\hspace{0.2in} \textbf{superior temporal} & 0.147 (0.014) & 0.138 (0.015) & 0.606 & \textbf{\textit{.0017}}\\
\hspace{0.2in} \textbf{inferior temporal} & 0.148 (0.016) & 0.139 (0.018) & 0.513 & \textbf{\textit{.0063}}\\
\hspace{0.2in} superior nasal & 0.190 (0.035) & 0.180 (0.034) & 0.281 & \textit{.0834}\\
\hspace{0.2in} inferior nasal & 0.165 (0.038) & 0.157 (0.032) & 0.227 & \textit{.1324}\\
\hspace{0.2in} \textbf{entire fundus} & 0.150 (0.013) & 0.142 (0.015) & 0.590 & \textbf{\textit{.0022}}\\
\hspace{0.2in} macula & 0.127 (0.017) & 0.120 (0.020) & 0.391 & \textit{.0278}\\
\\
\multicolumn{2}{l}{Vessel graph properties}\\
\hspace{0.2in} \textbf{number of nodes} & 371.04 (48.35) & 326.94 (41.29) & 0.981 & \textbf{\textit{$<$ .0001}}\\
\hspace{0.2in} \textbf{number of branches} & 378.62 (49.05) & 328.56 (44.16) & 1.073 & \textbf{\textit{$<$ .0001}}\\
\hspace{0.2in} \textbf{total length} & 11776.53 (846.25) & 11056.87 (978.21) & 0.787 & \textbf{\textit{$<$ .0001}}\\
\\
\textbf{FAZ normalized radius} & 0.118 (0.024) & 0.131 (0.022) & 0.554 & \textbf{\textit{.0036}}\\
\hline
\end{tabular}}
\end{center}
\end{table}

\subsection{Phase 3: Verification Results}

The verification results appear in Table~\ref{tab:test_results} in which the average and SD values for male and female groups, along with the effect sizes, p-values, and BH-adjusted p-values, are reported. Out of the nine parameters that resulted in significant sex differences on the exploration dataset, five again showed significant results in the expected direction on the verification dataset. Male fundus images showed more nodes and branches in the vessel graph as well as a greater total length of branches. Also, the area covered by the vessels in the superior temporal quadrant of the retina was higher in men. These results confirm that retinal vasculature is more prominent in males compared to females. However, no significant differences were observed in vessel coverage of the other quadrants of the retina and in the FAZ radius. Female fundi showed significantly brighter peripapillary areas aligned with what was observed in the CNN interpretation results. There was no significant difference in the normalized area of the optic disc.

\begin{table}[ht]
\renewcommand{\arraystretch}{1}
\caption[Hypotheses testing results]{Statistical test results for sex differences based on the verification dataset.}
\label{tab:test_results}
\begin{center}    
\resizebox{\textwidth}{!}{
\begin{tabular}{lcccr}
\hline 
\hline
Measurement & Male average (SD) & Females average (SD) & Effect size & BH-adjusted \textit{p-value}\\
\hline
\hline
\\
OD normalized area & 0.110 (0.022) & 0.107 (0.022) & 0.119 & \textit{.1770}\\
\\
\textbf{Peripapillary area brightness} & 0.669 (0.054) & 0.682 (0.054) & 0.243 &  \textbf{\textit{.0234}}\\
\\
\multicolumn{2}{l}{Vessel coverage}\\
\hspace{0.2in} \textbf{superior temporal} & 0.142 (0.019) & 0.139 (0.017) & 0.194 & \textbf{\textit{.0484}}\\
\hspace{0.2in} inferior temporal & 0.142 (0.021) & 0.143 (0.019) & 0.049 & \textit{.4039}\\
\hspace{0.2in} entire fundus & 0.145 (0.018) & 0.145 (0.017) & 0.007 & \textit{.4160}\\
\\
\multicolumn{2}{l}{Vessel graph}\\
\hspace{0.2in} \textbf{number of nodes} & 349.31 (57.51) & 332.66 (53.70) & 0.299 & \textbf{\textit{.0136}}\\
\hspace{0.2in} \textbf{number of branches} & 351.44 (62.82) & 334.98 (58.19) & 0.272 & \textbf{\textit{.0157}}\\
\hspace{0.2in} \textbf{total length} & 11381.20 (1173.34) & 11147.80 (1085.52) & 0.206 & \textbf{\textit{.0451}}\\
\\
FAZ normalized radius & 0.1295 (0.032) & 0.1288 (0.032) & 0.021 & \textit{.4160}\\
\hline
\hline
\end{tabular}}
\end{center}
\end{table}

\subsubsection{Mild Peripapillary Abnormalities}

\begin{figure}
\begin{center}
\resizebox{\textwidth}{!}{
\begin{tabular}{cccc}
\includegraphics[width=1.8in]{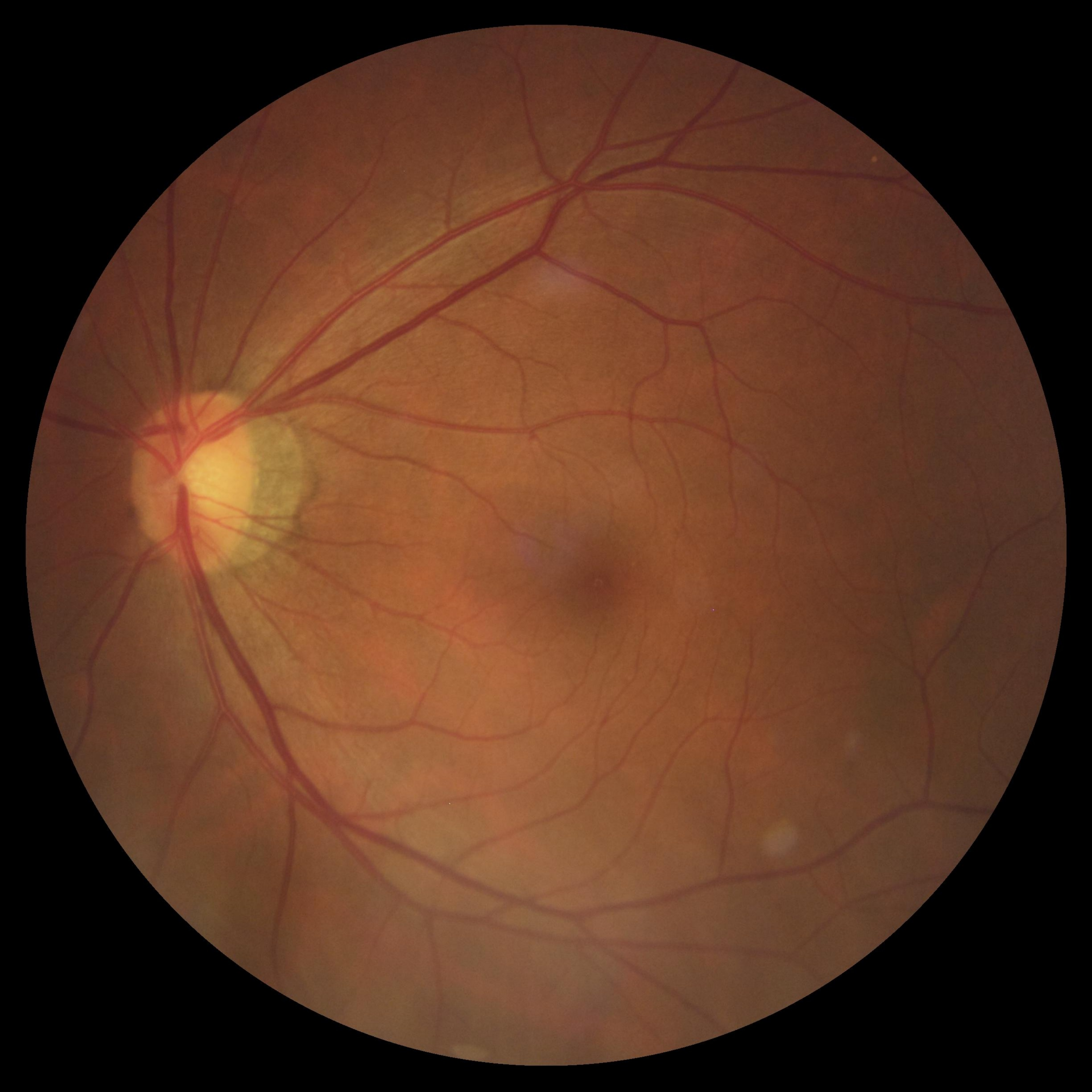} &  
\includegraphics[width=1.8in]{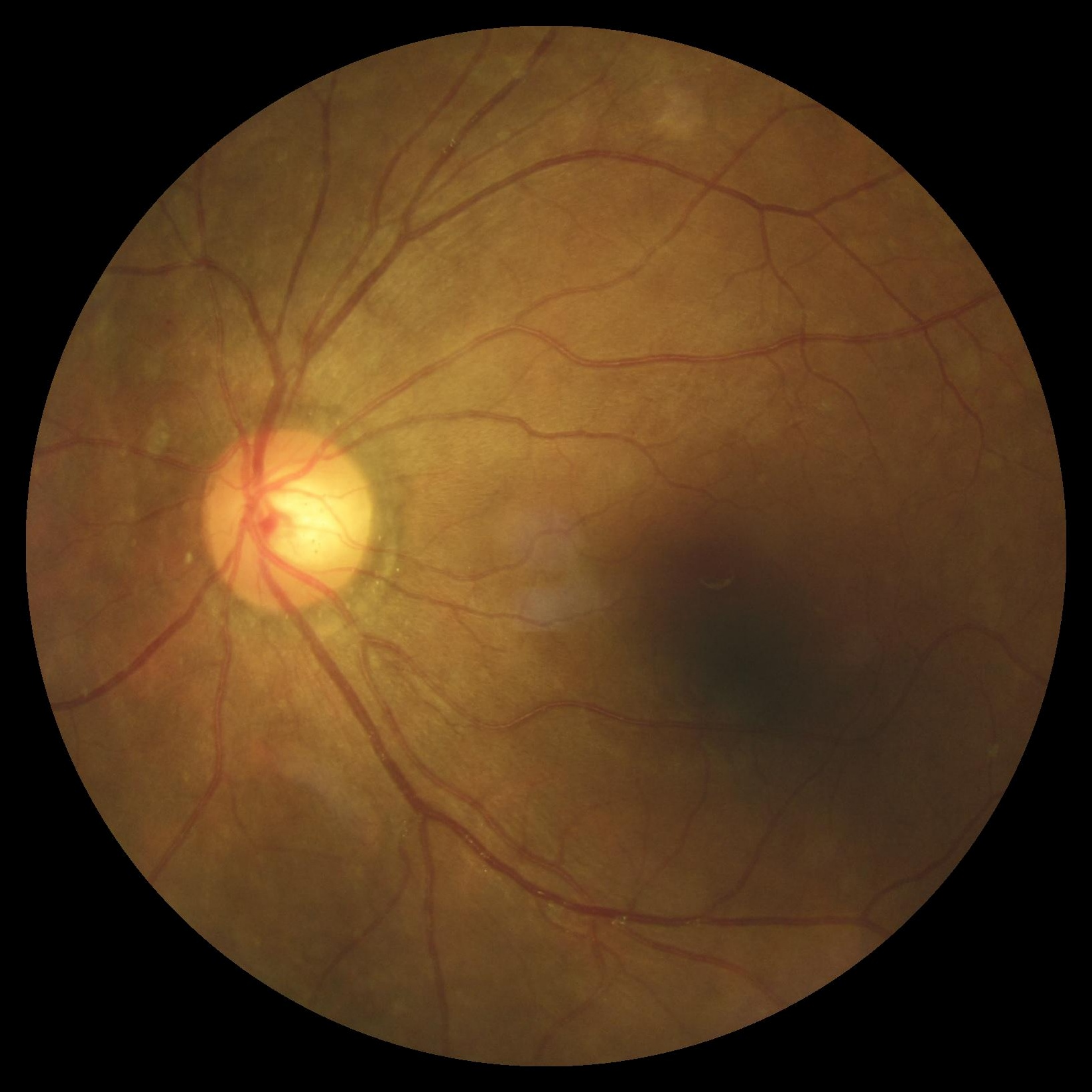} &
\includegraphics[width=1.8in]{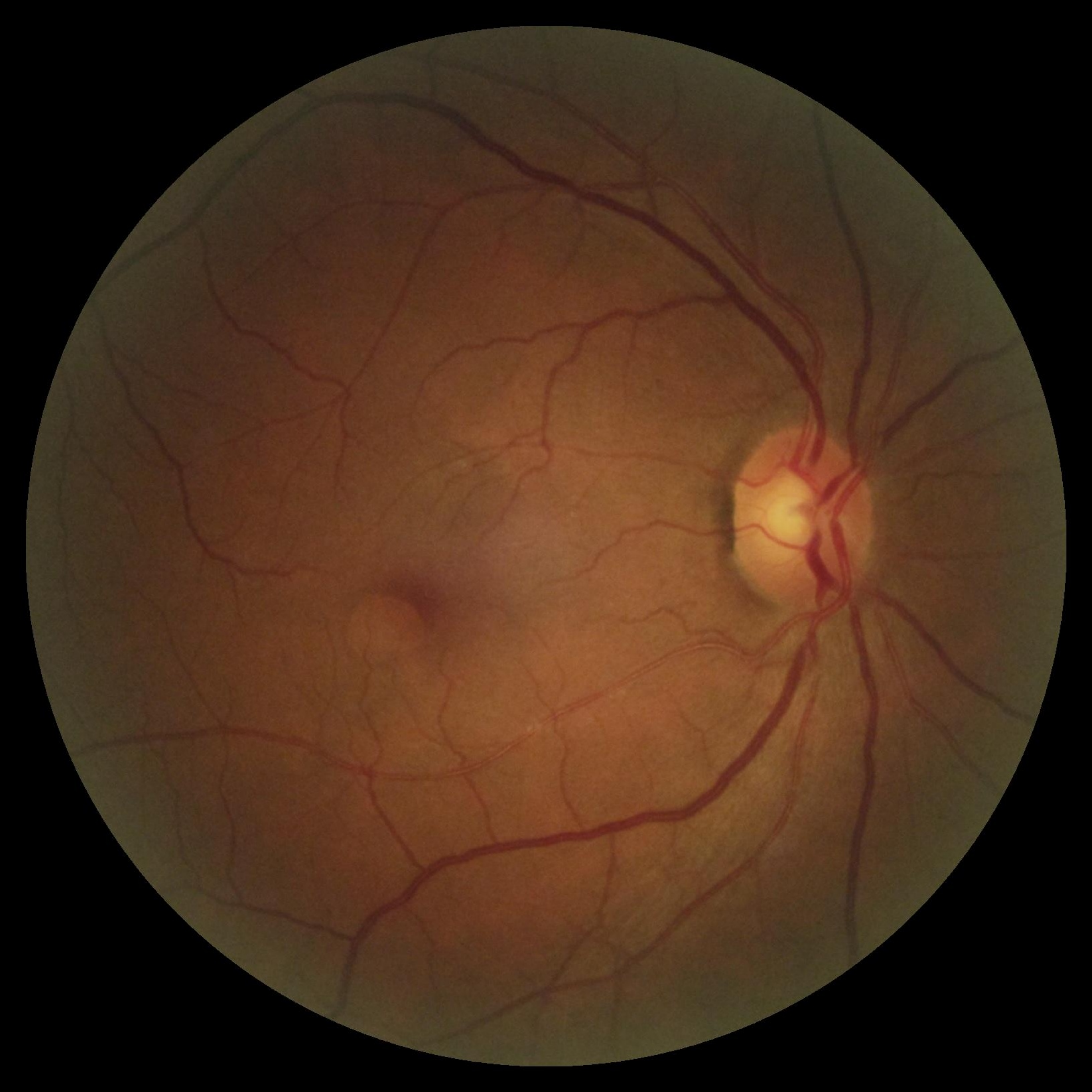} &  
\includegraphics[width=1.8in]{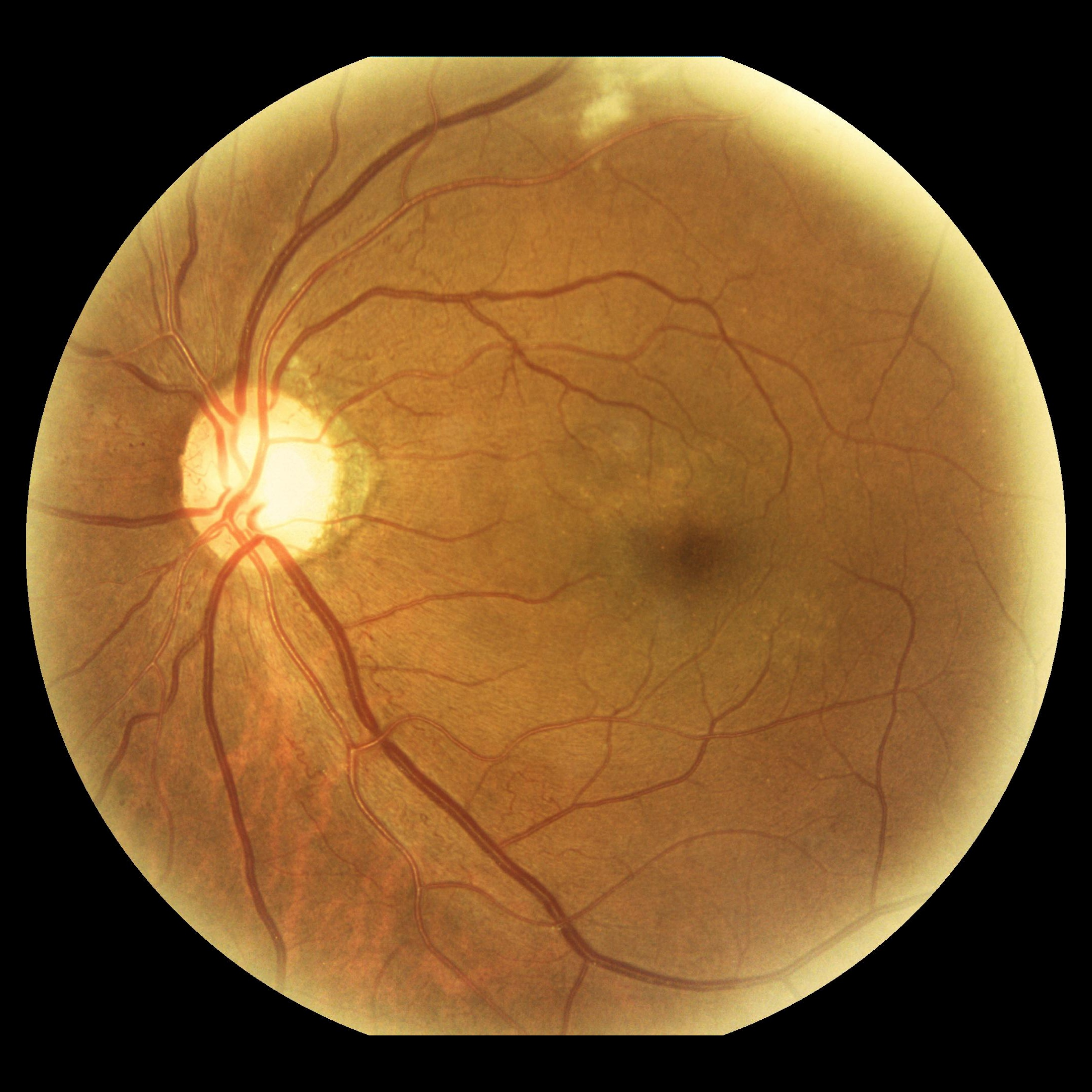} \\
(a) Tilted optic disc & (b) PPA &
(c) Pigmented crescent & (d) Scleral crescent\\
\end{tabular}}
\end{center}
\caption[Example peripapillary abnormalities]{An example fundus image from the ODIR dataset is shown for each of the peripapillary abnormalities.} 
\label{fig:example_abnormalities}
\end{figure}

\section{Discussion}

\begin{figure}
\begin{center}
\begin{tabular}{cc}
\includegraphics[width=3in]{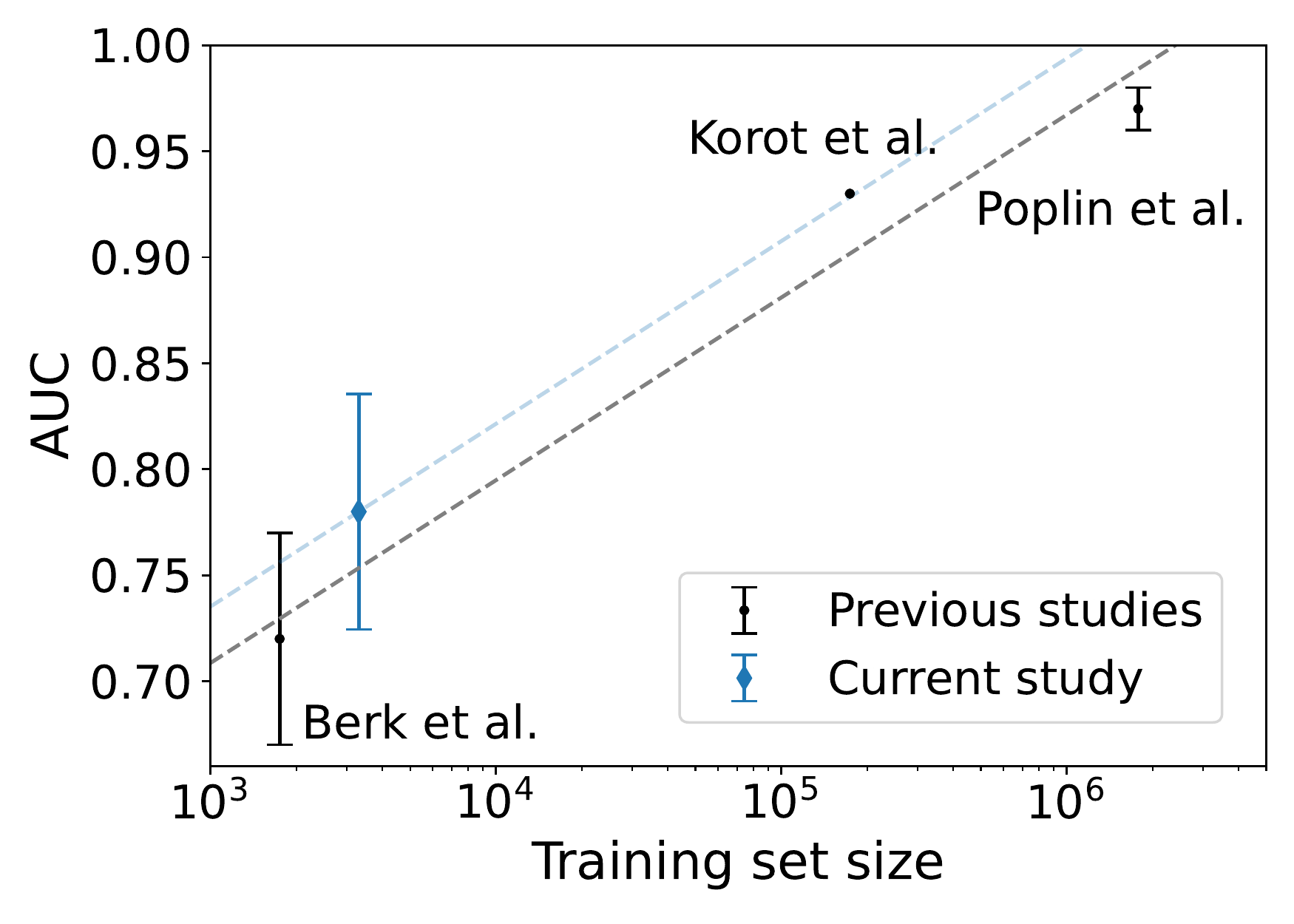} & \includegraphics[width=3in]{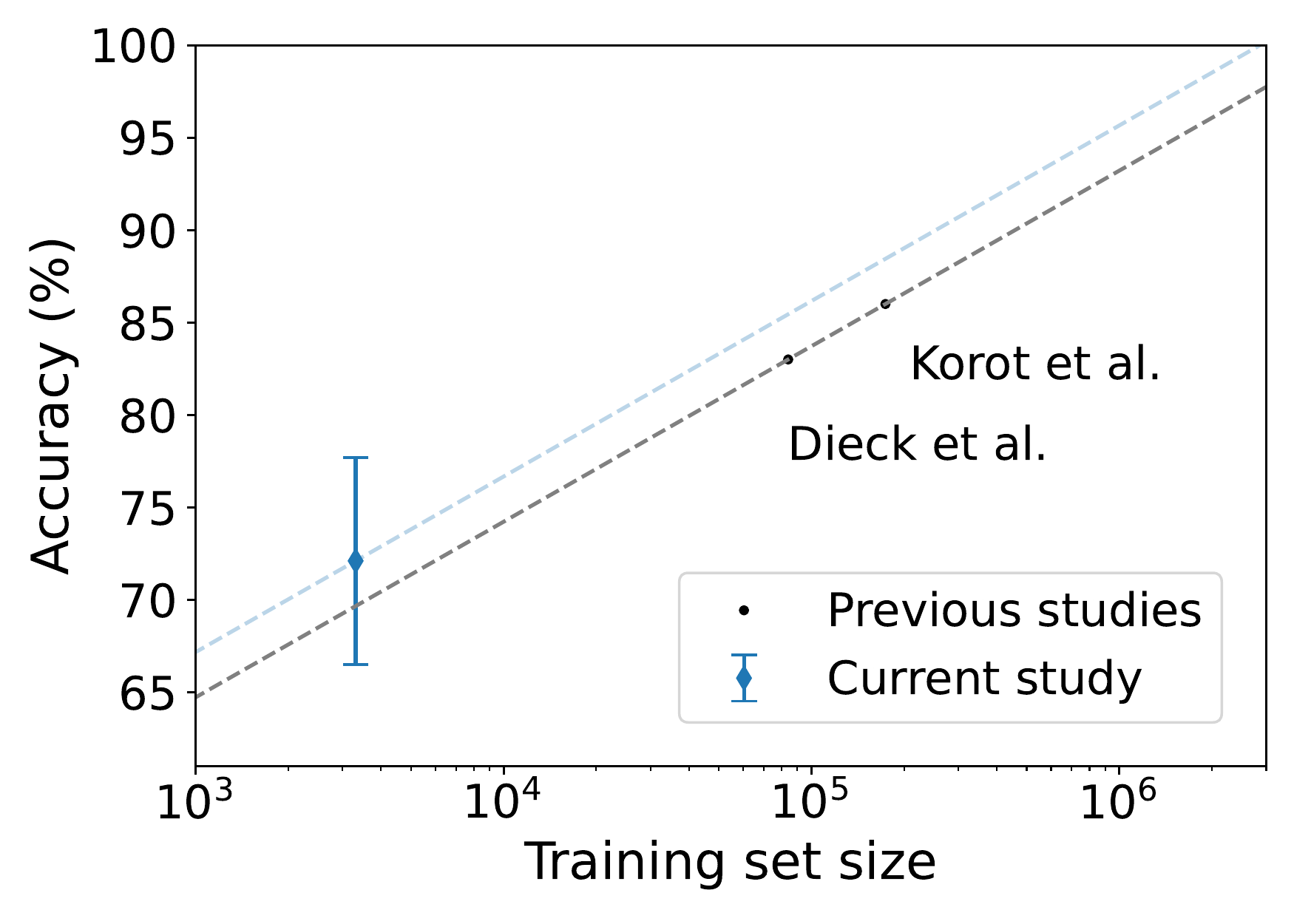}\\ 
(a)&
(b)\\
\end{tabular}
\end{center}
\caption[Result comparison with the previous studies]{The AUC and accuracy of the current study are compared to previous works in panels (a) and (b), respectively. The black line is the best line fitted to the previous results. Another line is plotted in blue with the same slope as the black line but passes through the results of the current work to simplify the comparison between studies with different training set sizes.} 
\label{fig:comparison} 
\end{figure}

The CNN models we trained to classify patient sex based on fundus images all achieved significantly above chance-level performance, confirming that this trait is visible to AI in contrast to the expert human eye. This suggests that there are features present in retinal fundus photographs that differ between males and females allowing the AI to predict sex reliably. Successful classification of patient sex has been reported in previous studies using various fundus datasets with different sizes \cite{poplin2018prediction, korot2021predicting, dieck2020factors, berk2022learning}. Machine learning performance relies heavily on the amount of data used for training, and this dependency is observed especially severely in the subcategory of deep neural networks. For example, the ImageNet \cite{deng2009imagenet} pretrained models used in the transfer learning framework are trained with about one million images. Considering this reliance on training dataset size, the moderate performance we achieved in the present study should be viewed in the context of the very small dataset we used for this purpose. The AUC and accuracy scores obtained by previous works are compared with the current study in Table~\ref{tab:comparison} and Fig.~\ref{fig:comparison}. In Fig.~\ref{fig:comparison}, the previous studies and the current study are denoted by black dots and blue diamonds, respectively, and error bars represent the confidence intervals. As shown, AUC score appears to have an approximately logarithmic relationship with the size of the training set (the x-axis is in logarithmic scale in Fig.~\ref{fig:comparison}). The gray dashed line shows the best fit line to the results of previous studies. The AUC of 0.79 achieved in the current work is placed above this line suggesting that the network is classifying sex better than the expected performance with this dataset size. Although Poplin et al.~\cite{poplin2018prediction} used a massive dataset with about 1.8 million images which is about 540 times larger than the training set in the current work, there is less than 20\% reduction in the AUC score. A study by Berk et al.~\cite{berk2022learning} provides a more commensurate comparison as they used similar datasets with similar sizes and obtained a 0.72 AUC score. It is worth mentioning that in Berk et al.~\cite{berk2022learning} this score was achieved and boosted by ensembling ten separately trained networks having a median AUC of 0.69. 

\begin{table}[ht]
\renewcommand{\arraystretch}{1.3}
\caption[Sex classification results of the previous studies]{Sex classification results of the previous studies and the current study.}
\label{tab:comparison}
\begin{center}       
\resizebox{0.6\textwidth}{!}{\begin{tabular}{lrrrr} 
\hline 
& Training set images & AUC & $CI_\alpha$ & Accuracy \\
\hline 
Poplin et al.~\cite{poplin2018prediction} & 1,779,020 & 0.97 & (0.96, 0.98) & - \\
Korot et al.~\cite{korot2021predicting} & 173,819 & 0.93 & - & 0.86 \\
Dieck et al.~\cite{dieck2020factors} & ~84,000 & - & - & 0.83 \\
\textbf{Current work} & \textbf{3,306} & \textbf{0.78} & \textbf{(0.73, 0.84)} & \textbf{0.72} \\ 
Berk et al.~\cite{berk2022learning} & 1,746 & 0.72 & (0.67, 0.77) & - \\ 
\hline 
\end{tabular}}
\end{center}
\end{table}

Next we discuss some of the sex differences revealed by the proposed methodology that are most likely used by AI to classify male and female fundus images. Male retinas were observed to have richer and more prominent retinal vasculature and also darker peripapillary areas compared to females. The prominence of retinal vessels can be linked to the higher metabolic demand in the male retina and, in turn, higher needs for energy and oxygen supply. In order to investigate possible alternative explanations for a darker peripapillary area in male retinal images, images with mild peripapillary abnormalities were identified. These abnormalities are subtle from the clinical point of view and are not considered a manifestation of any eye disease. Retinas with these abnormalities are also referred to as normal and healthy eyes. Dr.~Ozturan identified four types of mild peripapillary abnormalities in the images of the verification dataset, namely, tilted optic disc, peripapillary atrophy (PPA), pigmented crescent, and scleral crescent. For each of these conditions, a sample fundus image is shown in Fig.~\ref{fig:example_abnormalities}. Out of the 400 images in the verification dataset, 14 were identified with tilted optic disc, 75 with PPA, 36 with pigmented crescent, and 35 with scleral crescent. All images with any of these mild abnormalities were excluded from the verification dataset, and the brightness of the peripapillary area was again compared between males and females. Also, this parameter was compared between male and female eyes in each of these four conditions separately. Table~\ref{tab:abnormality} represents the resulting average, SD, effect size and p-value for each comparison. When images with any of the four types of mild peripapillary abnormalities were excluded, the finding of darker peripapillary region in males than in females persists. Although the p-value was pushed above the $\alpha=0.05$ ($\textit{p=.0529}$), there is no substantial decrease in the effect size compared to Table~\ref{tab:test_results}. The reduction observed in the p-values can be due to the lower number of samples in the latter test. Among the peripapillary abnormalities, only pigmented crescent group shows a significant sex difference in the brightness of the peripapillary area, which is again darker in males. Therefore, pigmented crescent might also be contributing to the lower p-value observed on the entire verification dataset. Also, based on the counts of the abnormality conditions Table~\ref{tab:abnormality}, there was no significant difference in the frequency of these peripapillary abnormalities between males and females based on a Chi-square test ($p=0.77$). Therefore, the difference observed in the brightness of the peripapillary area seems less likely to be because of the peripapillary abnormalities or their prevalence ratios.   

\begin{table}[ht]
\renewcommand{\arraystretch}{1.1}
\caption[Peripapillary abnormality results]{Sex differences in the peripapillary brightness for each of the mild peripapillary abnormality groups and for samples not belonging to any of theses groups.}
\label{tab:abnormality}
\begin{center} 
\resizebox{0.85\textwidth}{!}{
\begin{tabular}{c}
(a) Statistical test results.\\
\begin{tabular}{lcccr}
        \hline
        Condition & Male average (SD) & Females average (SD) & Effect size & \textit{p-value}\\[2pt]
        \hline
        Normal & 0.663 (0.050) & 0.673 (0.052) & 0.196 & \textit{.0529}\\
        Titled disc & 0.668 (0.039) & 0.685 (0.031) & 0.492 & \textit{.2055}\\
        PPA & 0.690 (0.065) & 0.706 (0.052) & 0.268 & \textit{.1268}\\
        Pigmented crescent & 0.640 (0.038) & 0.682 (0.053) & 0.911 & \textit{.0069}\\
        Scleral crescent & 0.729 (0.057) & 0.721 (0.063) & 0.471 & \textit{.0924}\\
        \hline
\end{tabular}\\
\\
(b) Counts of peripapillary abnormalities in VCH verification set.\\
\begin{tabular}{ccc}
        \hline
        Peripapillary abnormality & Male samples & Female samples \\[2pt]
        \hline
        Tilted disc & 7 & 7\\
        PPA & 31 & 44\\
        Pigmented crescent & 16 & 20\\
        Scleral crescent & 18 & 17\\
        \hline
\end{tabular}\\
\end{tabular}}
\end{center}
\end{table}

The findings are consistent with the previous studies applying deep learning models on retinal photographs to predict sex as they have also reported attention to the same anatomical parts of the retina \cite{poplin2018prediction, dieck2020factors, korot2021predicting, ilanchezian2021interpretable}. Although other measurements and parameters have shown to differ significantly between males and females, the sex differences discovered in this study represent those that are used by the CNN model since they are hypothesized solely based on the network itself without the help of anatomical or physiological sciences. 

The effect sizes are relatively small (ranging from 0.194 to 0.299), suggesting that the network uses multiple but subtle differences in the retina to classify patient's sex. The results of Ref.~\citenum{yamashita2020factors} also show a similar pattern by identifying the parameters that slightly differ between males and females. Given ophthalmologists are unable to see this trait in fundoscopic images, it is not surprising that the uncovered sex differences are relatively subtle. More prominent differences in the fundus images as a clue to patient's sex, would lead experts to identify them as in other relatively obvious traits such as age and ethnicity.

\section{Conclusions}

This study proposes a novel deep-learning-based approach in retinal fundus imaging as a promising pathway to scientific discovery. We have illustrated that AI is not only able to see and classify traits that are invisible to human experts but also can reveal previously unknown features in this imaging modality. Using patient sex as a case study, we have demonstrated that AI can be employed through interpretation techniques followed by statistical tests to discover anatomical and physiological differences in the retina. 

We predicted patient sex, a trait invisible to ophthalmologists in fundus imaging modality, significantly above chance level performance by fine-tuning a pre-trained CNN. The model showed a higher performance than would be expected based on previous studies considering the limited training set size. Furthermore, it was demonstrated that horizontal flip, the image transformation introduced in this study, was effective and improved both test AUC and accuracy. This data transformation technique is applicable in classifying the majority of eye diseases and other general traits where the label is irrelevant to whether the image is taken from the left or right eye.

We demonstrated the utility of two major CNN interpretation techniques, saliency maps and feature visualization, to hypothesize the possible sex differences in retinal fundus images. We observed that the model attends to the retina's main anatomical parts, namely, vessels, optic disc, and fovea, to classify males and females. By testing our a priori hypotheses, some of the sex differences in the retina were identified: Males showed a significantly darker peripapillary area and, in general, a more prominent retinal vasculature network. 

Future work will confirm further whether the discovered sex differences in the retinal fundus images can be incorporated into training of diagnosticians (e.g. ophthalmologists, optometrists) via psychophysical testing. However, the applications of the proposed methodology can go beyond patient sex and be extended to other traits and features of clinical importance. As the most promising example, relying on the ability of deep learning models to predict neurodegenerative diseases such as Alzheimer's Disease \cite{tian2021modular}, the same methodology can be used to uncover the manifestations and hallmarks visible in retinal fundus photographs to develop new diagnosis methods. 

In sum, this study demonstrates that the proposed three-phase methodology, which is solely based on AI techniques, leads to new anatomically translatable and physiologically plausible discoveries that can be applied to various applications.

\section{Disclosures}

The authors declare that there is no conflict of interest.

\section{Acknowledgements}

This work was supported by a Natural Sciences and Engineering Research Council of Canada Discovery Grants RGPIN-2019-05554 (IO) and an Accelerator Supplement RGPAS-2019-00026 (IO), a Canada Foundation for Innovation, John R. Evans Leaders Fund (IO), UBC Data Science Institute postdoctoral award (GO, IO \& OY), UBC DMCBH Kickstart award (IO \& OY), and a UBC Health Innovation Funding Investment award (IO \& OY), an NSERC Discovery Grant (22R82411) (OY). OY also acknowledges support by the Pacific Institute for the Mathematical Sciences (PIMS). This research was also supported in part through computational resources and services provided by Advanced Research Computing at the University of British Columbia \cite{ARC}.


\bibliography{article}   
\bibliographystyle{spiejour}   


\listoffigures
\listoftables
\pagebreak

\newcommand{\beginsupplement}{%
        \setcounter{section}{0}
        \renewcommand{\thesection}{S\arabic{section}}%
        \setcounter{table}{0}
        \renewcommand{\thetable}{S\arabic{table}}%
        \setcounter{figure}{0}
        \renewcommand{\thefigure}{S\arabic{figure}}%
     }

\beginsupplement

\section{Vessel Segmentation Details}

\subsection{DRIVE dataset}
\begin{figure}[!ht]
    \centering
    \includegraphics[width=3.33in]{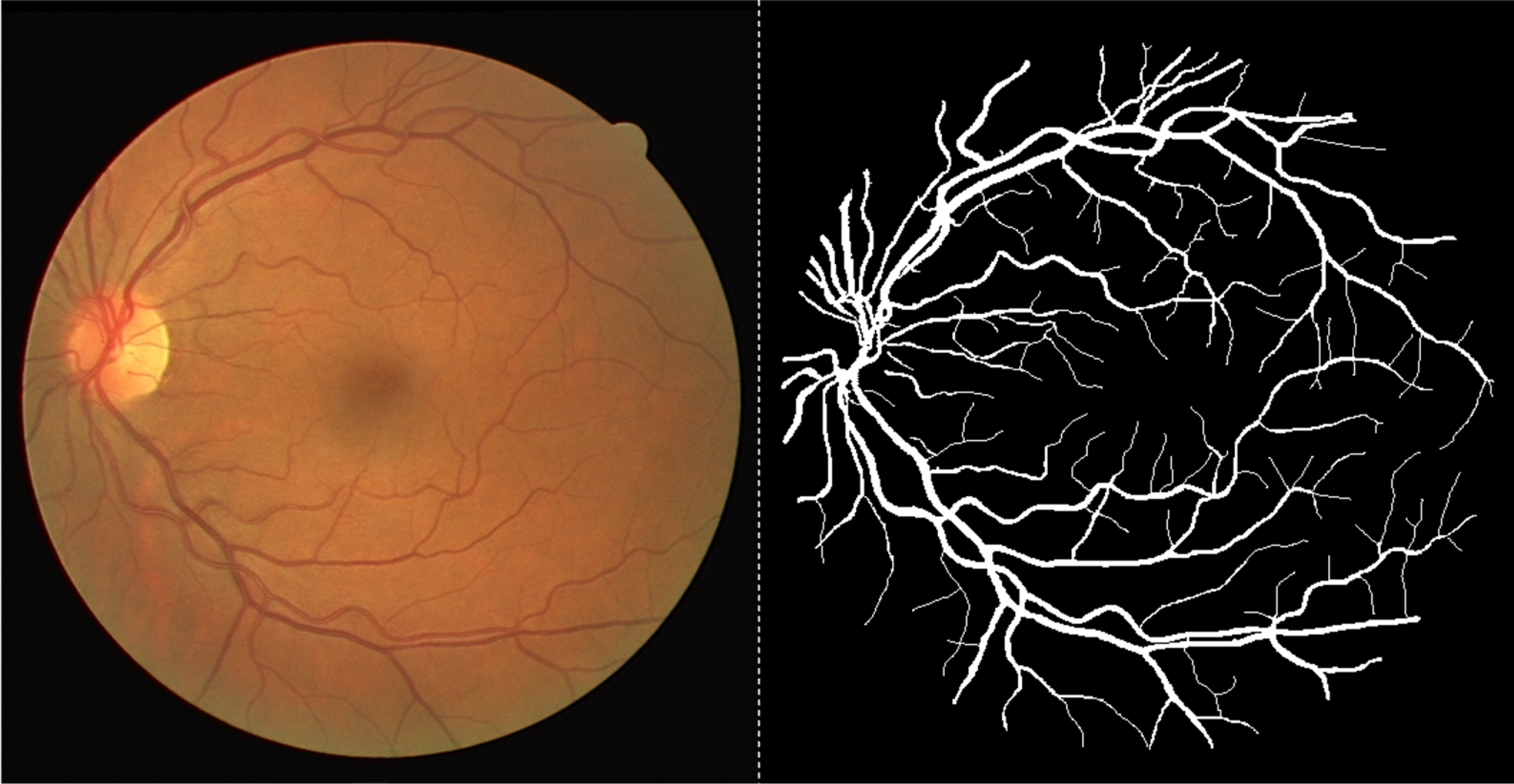}
    \caption[A sample from DRIVE dataset]{A sample pair of a fundus image and its binary vessel mask from DRIVE dataset.}
    \label{fig:drive} 
\end{figure}

DRIVE \cite{staal:2004-855} is one of the most used benchmark datasets for retinal vessel segmentation, and numerous papers have published the performance of their automated vessel segmentation method on this dataset. DRIVE contains 40 samples which are originally separated into two sets of equal size of 20 as the training and test sets. Each sample is a pair of a fundus image and a black-and-white binary mask of the retinal vessels as shown Fig.~\ref{fig:drive}. 

\subsection{Training procedure}

\begin{figure}
\begin{center}
\begin{tabular}{c}
\includegraphics[width=3in]{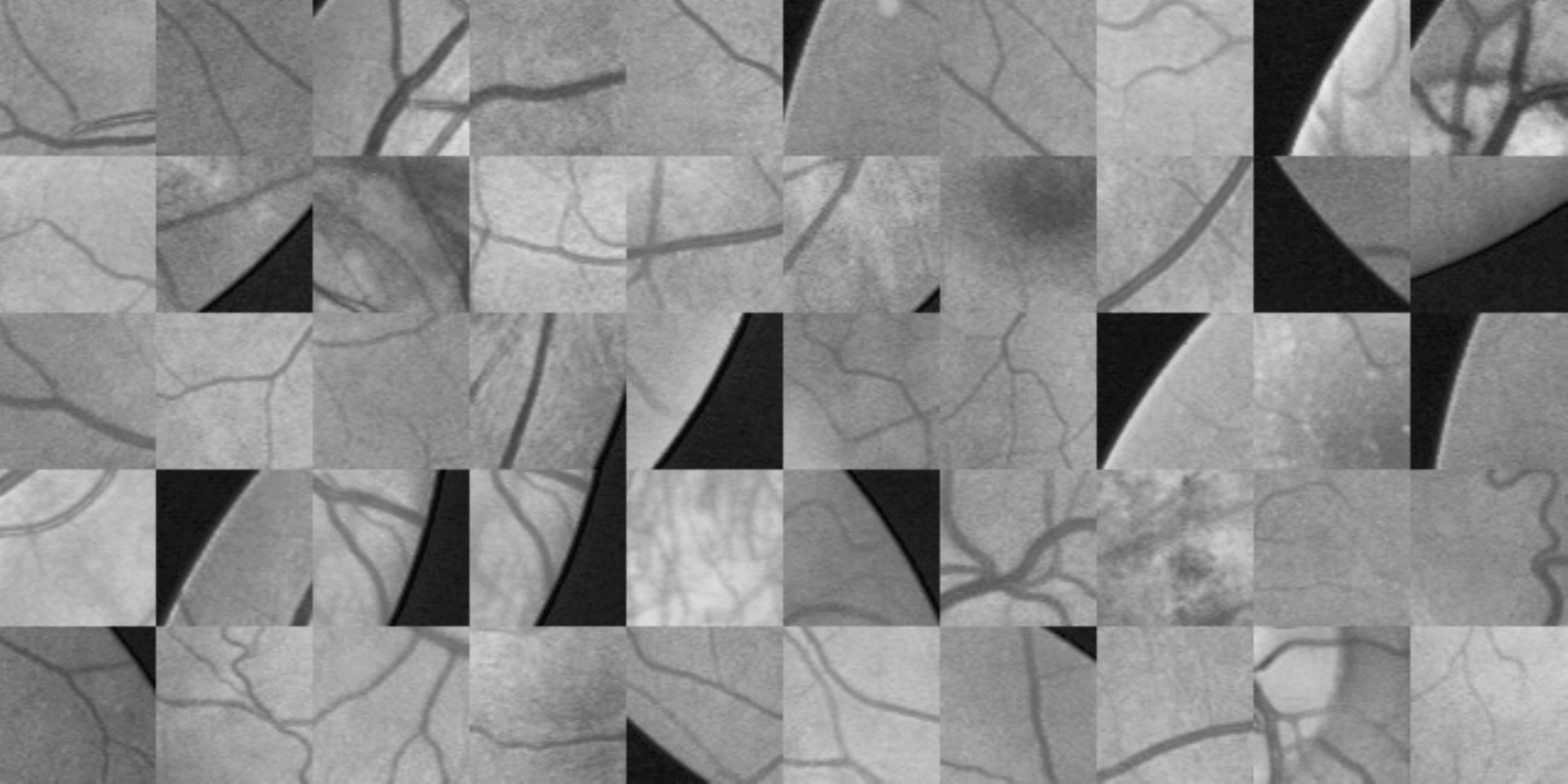}\\ 
(a) image patches\\
\\
\includegraphics[width=3in]{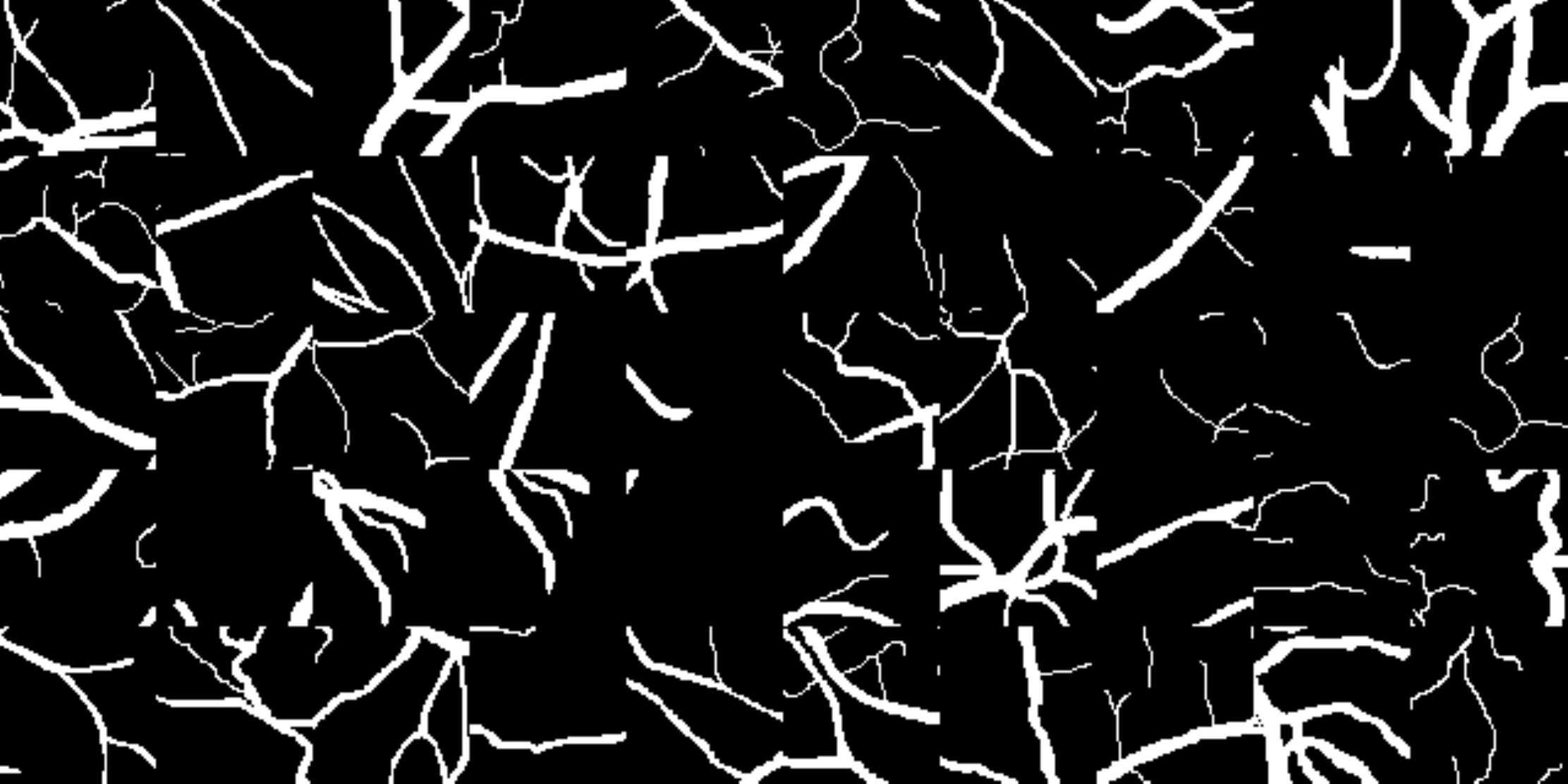} \\
(b) vessel masks\\
\end{tabular}
\end{center}
\caption[Sample DRIVE patches]{Sample training patches and vessel masks obtained from DRVIE dataset.} 
\label{fig:sample_input} 
\end{figure}

Based on Ref.~\citenum{lee_zq_2021} implementation, images were cropped into multiple smaller and overlapping patches to enhance the dataset size as it has a limited number of images. The patch size was set to 64 and the stride to 16, which resulted in 150,000 image patches. Some sample training patches, along with their ground truth vessel masks, are shown in Fig.~\ref{fig:sample_input}. The hyperparameters are summarized in Table~\ref{tab:hyperparameters_vesseg}.

Then, to take advantage of the entire DRIVE dataset and enhance the performance, the same parameters were used to train the model using both training and test sets. The resulting model was applied to the VCH dataset for obtaining final vessel segmentation masks. Also, because of the domain adaptation issues, the VCH images were first preprocessed before being fed to the model. The preprocessing procedure involved increasing contrast by a factor of 1.5, decreasing brightness by a factor of 0.6, and slightly blurring the images with a Gaussian blur filter. These steps were taken to make VCH images get closer to DRIVE images that the model was trained on, and results were assessed qualitatively by try and error to choose the best preprocessing hyperparameters.

\begin{figure}
    \centering
    \makebox[\textwidth]{\includegraphics[width=1\textwidth]{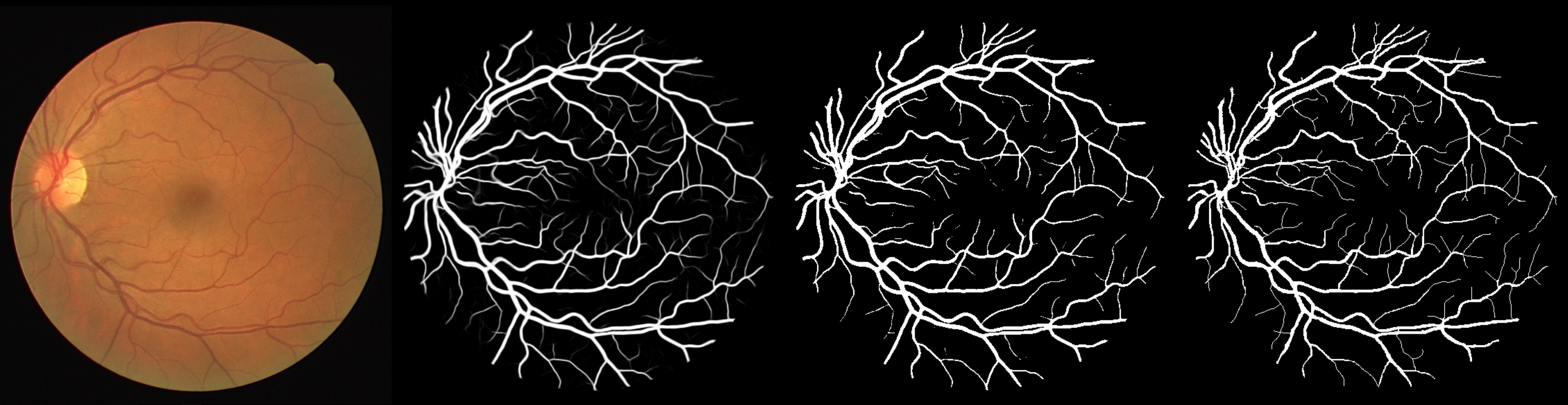}}
    \caption[Sample vessel segmentation output]{A sample vessel segmentation output based on the DRIVE test set. From left to right, the original fundus image, the output vessel probability mask, the output vessel mask (the probability mask thresholded at 0.5), and the ground truth vessel mask.}
    \label{fig:vesseg_sample} 
\end{figure}

After training, the model achieved a test AUC score of 0.987 and an F1 score of 0.853. Fig.~\ref{fig:vesseg_sample} shows sample vessel segmentation results from the test set.

\begin{table}[!ht]
\renewcommand{\arraystretch}{1.2}
\caption[Vessel segmentation hyperparameters]{Summary of the hyperparameters for training the vessel segmentation model.}
\label{tab:hyperparameters_vesseg}
\begin{center}    
\resizebox{0.35\textwidth}{!}{
\begin{tabular}{lrr} 
\hline 
\hline
\multicolumn{3}{c}{\textbf{\large Hyperparameters}} \\[5pt]
\multicolumn{3}{l}{\textbf{Data}}\\[2pt]
\hline
& \textbf{train patch height} & 64  \\
& \textbf{train patch width} & 64  \\
& \textbf{test patch height} & 64  \\
& \textbf{test patch width} & 64  \\
& \textbf{stride height} & 16  \\
& \textbf{stride width} & 16  \\
& \textbf{number of patches} & 150,000  \\
& \textbf{validation ratio} & 0.1  \\
\\
\multicolumn{3}{l}{\textbf{Training}}\\[2pt]
\hline
& \textbf{optimizer} & Adam \\
& \textbf{loss} & binary cross-entropy \\
& \textbf{learning rate} & 0.0005 \\
& \textbf{batch size} & 16  \\
& \textbf{stride height} & 16  \\
& \textbf{number of epochs} & 20  \\
\\
\multicolumn{3}{l}{\textbf{Network}}\\[2pt]
\hline
& \textbf{architecture} & LadderNet \\
& \textbf{number of layers} & 3 \\
& \textbf{number of filters} & 16 \\
\\
\hline 
\hline
\end{tabular}}
\end{center}
\end{table}

\end{spacing}
\end{document}